\documentclass[preprint,review,12pt,authoryear]{elsarticle}
\usepackage{amssymb, amsmath, psfrag, graphicx,color}
\usepackage{units,booktabs,url,makecell,xcolor,multirow, diagbox}
\definecolor{darkpastelgreen}{rgb}{0.01, 0.75, 0.24}

\usepackage[T1]{fontenc}
\usepackage[hidelinks]{hyperref}
\usepackage{tikz}
\usetikzlibrary{calc,fadings,arrows,shapes,backgrounds}
\usepackage{wrapfig}
\usepackage{subcaption}
\usepackage[prefix]{nomencl}
\makenomenclature
\usepackage{etoolbox}
\renewcommand\nomgroup[1]{%
	\item[\bfseries
	\ifstrequal{#1}{A}{\hspace{-.01cm}List of Acronyms}{
	\ifstrequal{#1}{S}{\hspace{-.15cm}List of Symbols}{}}
]}
\usepackage{epstopdf}
\newcommand{\wrt}{\text{w.r.t. }}

\journal{Computers \& Chemical Engineering}

\begin{document}
	
\begin{frontmatter}
\title{Data-based Design of Inferential Sensors for Petrochemical Industry}
\author[STUBA,cor1]{Martin Mojto}
\author[SLOVNAFT]{Karol \v Lubu\v sk\'y}
\author[STUBA]{Miroslav Fikar}
\author[STUBA]{Radoslav Paulen}
\cortext[cor1]{Tel.: +421 (0)2 5932 5349, Mail: martin.mojto@stuba.sk (M. Mojto)}
\address[STUBA]{Faculty of Chemical and Food Technology, Slovak University of Technology in Bratislava, Radlinsk\'eho 9, 812 37 Bratislava, Slovakia}
\address[SLOVNAFT]{Slovnaft, a.s., Vl\v cie hrdlo 1, 824 12 Bratislava, Slovakia}

\begin{abstract}
	Inferential (or soft) sensors are used in industry to infer the values of imprecisely and rarely measured (or completely unmeasured) variables from variables measured online (e.g., pressures, temperatures). The main challenge, akin to classical model overfitting, in designing an effective inferential sensor is the selection of a correct structure of the sensor. The sensor structure is represented by the number of inputs to the sensor, which correspond to the variables measured online and their (simple) combinations. This work is focused on the design of inferential sensors for product composition of an industrial distillation column in two oil refinery units, a Fluid Catalytic Cracking unit and a Vacuum Gasoil Hydrogenation unit. As the first design step, we use several well-known data pre-treatment (gross error detection) methods and compare the ability of these approaches to indicate systematic errors and outliers in the available industrial data. We then study effectiveness of various methods for design of the inferential sensors taking into account the complexity and accuracy of the resulting model. The effectiveness analysis indicates that the improvements achieved over the current inferential sensors are up to 19\,\%.
\end{abstract}

\end{frontmatter}

\printnomenclature
\newpage

\section{Introduction}
\nomenclature[A, 01]{AIC$_\text{C}$}{Corrected Akaike Information Criterion}
\nomenclature[A, 02]{BC}{Bias Correction}
\nomenclature[A, 03]{BIC}{Bayesian Information Criterion}
\nomenclature[A, 04]{CV}{Controlled Variable}
\nomenclature[A, 05]{FCC}{Fluid Catalytic Cracking}
\nomenclature[A, 06]{GF}{Gasoline Fraction}
\nomenclature[A, 07]{HGO}{Hydrogenated Gasoil}
\nomenclature[A, 08]{LASSO}{Least Absolute Shrinkage and Selection Operator}
\nomenclature[A, 09]{MCD}{Minimum Covariance Determinant}
\nomenclature[A, 10]{MIQP}{Mixed-Integer Quadratic Programming}
\nomenclature[A, 11]{NIPALS}{Nonlinear Iterative Partial Least Squares}
\nomenclature[A, 12]{OLSR}{Ordinary Least Squares Regression}
\nomenclature[A, 13]{PCA}{Principal Component Analysis}
\nomenclature[A, 14]{PLS}{Partial Least Squares}
\nomenclature[A, 15]{Ref}{Current (reference) inferential sensor}
\nomenclature[A, 16]{RMSE}{Root Mean Square Error}
\nomenclature[A, 17]{RSS}{Residual Sum of Squares}
\nomenclature[A, 18]{SS}{Subset Selection}
\nomenclature[A, 19]{SS-CV}{Subset Selection with Cross-Validation}
\nomenclature[A, 20]{SVD}{Singular Value Decomposition}
\nomenclature[A, 21]{VGH}{Vacuum Gasoil Hydrogenation}
\nomenclature[S, 01]{$a$}{Vector of inferential sensor parameters, $a\in\mathbb R^{n_{\text{p}}}$}
\nomenclature[S, 02]{$F$}{Flow rate}
\nomenclature[S, 03]{$H_v$}{Heat of vaporization}
\nomenclature[S, 04]{$n$}{Number of measurements}
\nomenclature[S, 05]{$n_{\text{p}}$}{Total number of available inputs}
\nomenclature[S, 05]{$n_{\text{p}}^*$}{Number of inputs selected for the sensor structure}
\nomenclature[S, 05]{$\tilde n_{\text{p}}$}{Number of principal components}
\nomenclature[S, 06]{$m$}{Vector of input variables, $\mathbb R^{n_{\text{p}}}$}
\nomenclature[S, 07]{$M$}{Matrix of input dataset, $\mathbb R^{n\times n_{\text{p}}}$}
\nomenclature[S, 08]{$M_{\text{C}}$}{Matrix of input dataset centered, $\mathbb R^{n\times n_{\text{p}}}$}
\nomenclature[S, 09]{$M_{\text{N}}$}{Matrix of input dataset normalized, $\mathbb R^{n\times n_{\text{p}}}$}
\nomenclature[S, 10]{$p$}{Pressure}
\nomenclature[S, 11]{$PCT$}{Pressure-compensated temperature}
\nomenclature[S, 12]{$Q$}{Energy flow rate}
\nomenclature[S, 13]{$R$}{Universal gas constant}
\nomenclature[S, 14]{$RX$}{Gas/liquid phase ratio}
\nomenclature[S, 15]{$T$}{Termodynamic temperature}
\nomenclature[S, 16]{$x$}{Concentration}
\nomenclature[S, 17]{$y$}{Vector of measurements of output variable, $\mathbb R^n$}
\nomenclature[S, 18]{$\hat{y}$}{Inferred CV by the inferential sensor}
\nomenclature[S, 19]{$y_{\text{C}}$}{Vector of measurements of output variable centered, $\mathbb R^n$}
\nomenclature[S, 20]{$y_{\text{N}}$}{Vector of measurements of output variable normalized, $\mathbb R^n$}
The accuracy and reliability of industrial measurements have a huge impact on the effectiveness of industrial process control~\citep{khatibisepehr_2013}. Especially, the control performance of advanced process controllers ~\citep{qin_2003} is highly related to the indication quality of controlled variables (CVs). It is often the case that the crucial CVs (e.g., distillate purity) are too expensive or impossible to measure at the frequency required for an effective feedback control. This gave rise to a use of so-called inferential (or soft) sensors~\citep{mejdell_1991, kordon_2003, curreri_2020}.

The purpose of an inferential sensor is to infer the CV value (output) using the data from other measured variables (inputs). The design procedure aims at a) identifying the sensor structure and b) at estimating the sensor parameters. While the latter problem can be solved relatively easily and there are standard academic and industrial tools even for situations of parameter-varying sensors (e.g., recursive estimation or simple bias update~\citep{king_2011}), the former issue of structure selection can be much more challenging in practice.

The effectiveness and reliability of inferential sensors are highly related to the quality of data used for the design. Subsequently, the data quality is affected by the amount of systematic and random errors~\citep{su_2009}. There are several methods dealing with the pre-treatment methods for industrial data~\citep{alves_2007}. The group of well-known and popular multivariate data treatment methods includes Hotelling's $T^2$ distance~\citep{hotelling_1931}, $k$-means clustering~\citep{forgy_1965}, or minimum covariance determinant (MCD) technique~\citep{rousseeuw_1984}. Several applications of these methods were reported in the industrial context~\citep{alameddine_2010, xu_2017, frumosu_2019, azzaoui_2019, fontes_2021}. 

There are several approaches to the design of inferential sensors~\citep{fortuna_2007, liu_2010, sun_2021}. According to the way of inferential sensor modeling, one can divide these methods into two main types: model-based and data-driven. The former category usually uses a first-principles model~\citep{torgashov_2019} and therefore it requires fundamental knowledge about the process behavior and characteristics. Several contributions have been published in the field of model-based approaches and their combination with the extended Kalman filter~\citep{gryzlov_2013} or with the neural networks~\citep{chen_2000}. However, the behavior of industrial process is often too complicated and requires much effort to develop a first-principles model with an acceptable accuracy. In such a case, data-driven approaches provide less demanding yet effective solution. The popularity of the data-driven methods also increases in the process industry because of the increased availability of modern and cheap online sensors.

Currently, the most popular data-based methods for inferential sensor design are based on Principal Component Analysis (PCA) regression~\citep{pearson_1901} and on Partial Least Squares (PLS)~\citep{wold_1984, wold_2001}. The principle of PCA regression is an application of unsupervised learning to input-variable space reduction and subsequent regression on the reduced space. The use of PCA has a long history yet its use is still very frequent in industry~\citep{kadlec_2009, yuan_2015, yu_2020}. The characteristics of PLS are similar to PCA regression~\citep{dunn_1989}, yet unlike PCA it takes into account also the output space (supervised learning approach). The selection between the use of PCA or PLS is dependent on the availability and quality of infrequently measured output variables.

Both PCA and PLS partially avoid overfitting of the inferential sensor by performing regression in the reduced dimensions. The structure of the resulting sensor is, however, not sparse which might be undesirable or prohibitive, e.g., in case of using the designed inferential sensor for advanced process control. As a response, sparsifying data-driven approaches for the soft-sensor design were developed. The Least Absolute Shrinkage and Selection Operator (LASSO)~\citep{santosa_1986, tibshirani_2011} uses 1-norm penalization balance between the soft-sensor accuracy and its complexity. The concept of sparse soft-sensor design is further developed in so-called subset selection (SS) methods, which aim at selecting the best subset of explanatory variables from the multivariate input space. The original methodology was proposed to select suitable input variables from the whole set of input candidates according to various (backward, forward, bi-directional) stepwise approaches~\citep{efroymson_1960, smith_2018}. Several studies~\citep{miyashiro_2015, mencarelli_2020} proved that SS can be enhanced by using model-overfitting criteria such as adjusted $\mathit{R}^2$ ($\mathit{R}^2_{\text{adj}}$), corrected Akaike Information Criterion (AIC$_\text{C}$), or Bayesian Information Criterion (BIC). The performance of SS can also be improved by emulation of the cross-validation process~\citep{takano_2020}. 

{In this work, we deal with the data treatment and with the subsequent design of inferential sensors on the pre-treated data. The purpose of the data treatment is to remove the outliers and systematic errors from measurements to make the design of inferential sensors more accurate and reliable. We compare the effectiveness of Hotelling's $T^2$ distance, MCD and $k$-means clustering to indicate outliers in a multivariate industrial dataset. Our methodology uses only data-based treatment while the model-based techniques such as data reconciliation also have the potential to enhance the quality of the final designed soft sensors~\citep{manenti_2011, xenos_2014}. The development of first-principles models for the case studies presented in this contribution would be rather complex and is often cumbersome in real industrial conditions. Therefore the data reconciliation is not considered in this paper.}

We design linear inferential sensors using various data-driven techniques. The studied methods involve variance-covariance approaches (PCA and PLS) and relatively recent model-sparsity enforcing methods (LASSO and SS). The main contribution of this study is the comparison of these methods in the context of industrial (inferential) soft-sensor design. We analyze the effectiveness of these methods investigating a soft-sensor performance in two industrial use cases provided by the refinery Slovnaft, a.s. in Bratislava, Slovakia. The examples differ in complexity yet they both aim at monitoring the product composition of a distillation column in a crucial processing unit. Recently, there have been several publications~\citep{demorais_2019, humod_2020, luo_2020} dealing with the design of inferential sensors for similar industrial units. The soft sensors presented in these publications show satisfying performance in the particular petrochemical process or part of the refinery. While these contributions are focused on a specific soft-sensor design method, we analyze the performance of several methods based on different principles.

The structure of the paper is organized as follows. At first, the
basic description of industrial use cases is introduced. Subsequently,
the key aspects and relations of well-known data treatment methods are
reviewed. Next, the advantages and important characteristics of
soft-sensor design methods are briefly introduced. Case studies
present results and compare obtained soft sensors for a Fluid Catalytic Cracking (FCC) unit and a Vacuum Gasoil Hydrogenation (VGH)
unit. The obtained results are finally discussed and the paper is concluded.

\section{Problem Description}
Our goal is to identify models of inferential sensors in the following linear form:
\begin{align}
	\hat{y}_i = m_i^\intercal(a_1, a_2, \ldots, a_{n_{\text{p}}})^\intercal = m_i^\intercal a,
\end{align}
where $\hat{y}$ stands for the desired CV inferred (estimated) by the sensor, $m$ is the vector of available input variables, $a\in\mathbb R^{n_\text{p}}$ represents the vector of sensor parameters, and index $i$ represents measurements index.

\subsection{FCC unit}
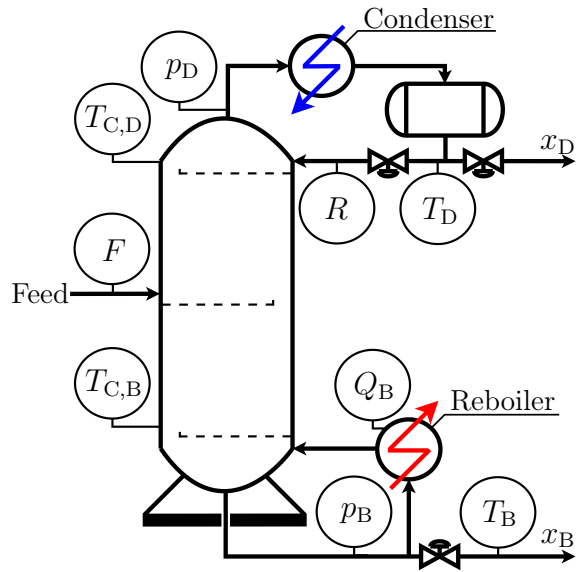
\begin{figure}
	\centering
	\begin{tikzpicture}[thick,scale=1, every node/.style={scale=1}]
		\node[anchor=south west,inner sep=0] (image) at (0,0) {\includegraphics[width=0.5\linewidth]{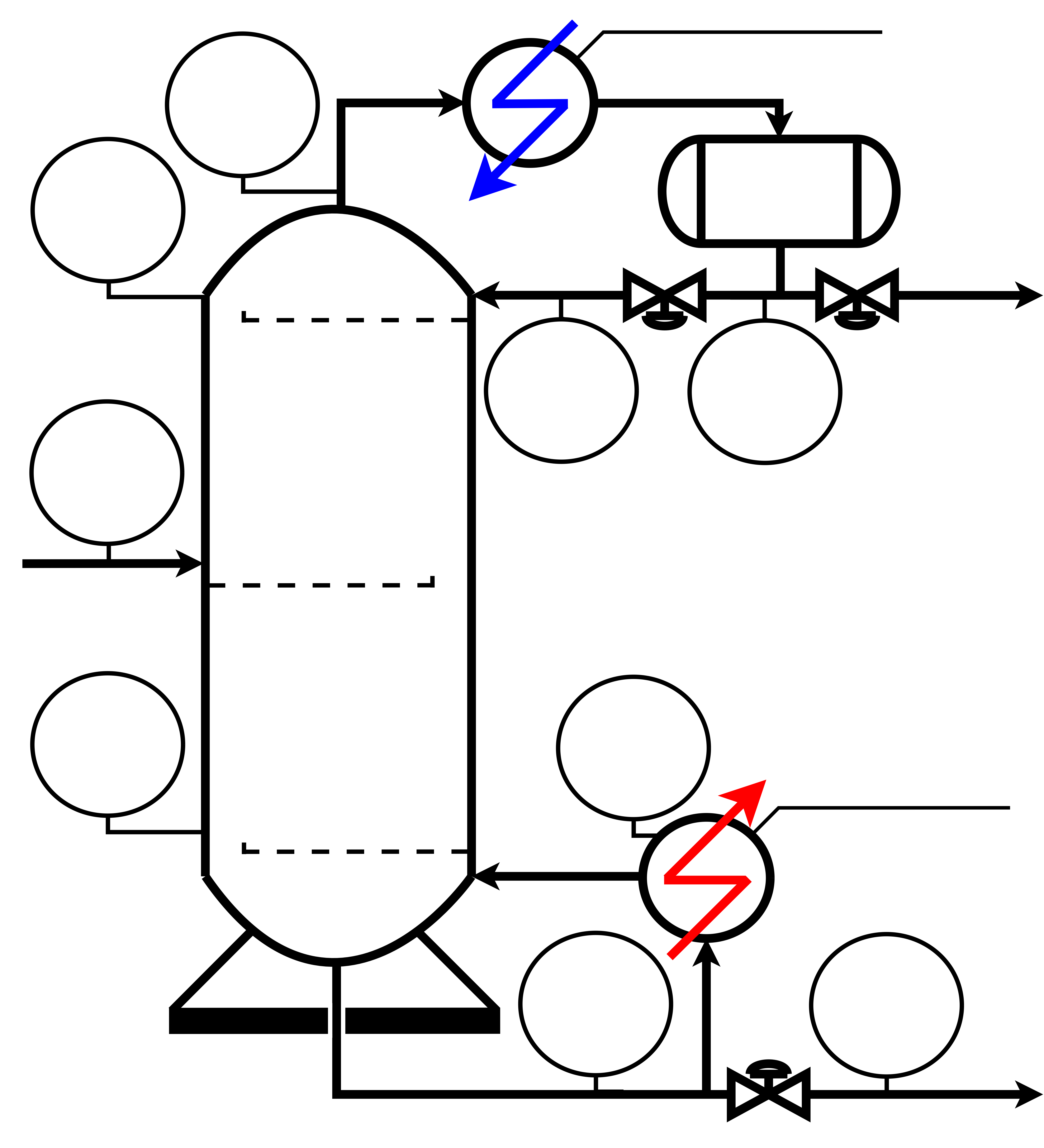}};
		\node at (140pt,214pt) {\small Condenser};
		\node at (168pt,69.2pt) {\small Reboiler};
		\node at (-7pt,110pt) {\small Feed};
		\node at (189pt,166pt) {$x_\text{D}$};
		\node at (189pt,17pt) {$x_\text{B}$};
		\node at (47pt,194.5pt) {$p_\text{D}$};
		\node at (113pt,26.5pt) {$p_\text{B}$};
		\node at (119.5pt,74.5pt) {$Q_\text{B}$};
		\node at (144.5pt,141pt) {$T_\text{D}$};
		\node at (167pt,26.5pt) {$T_\text{B}$};
		\node at (21pt,175pt) {$T_\text{C,D}$};
		\node at (21pt,75pt) {$T_\text{C,B}$};
		\node at (105pt,142pt) {$R$};
		\node at (21pt,126pt) {$F$};
	\end{tikzpicture}
	\caption{A schematic diagram of the depropanizer column.}
	\label{fig:dep_plant_scheme}
\end{figure}

This unit serves to convert heavy hydrocarbon fractions (vacuum distillates) of the crude oil incoming from the entire refinery to more valuable products, such as gasoline or olefins. The FCC unit is separated into several individual sections (sub-units). One of these sub-units includes several interconnected distillation columns (e.g., debutanizer or depropanizer) to process light hydrocarbons C2--C6. The desired variable ($\hat{y}$) to be inferred by the soft sensor is the composition (main impurity) of the bottom product $x_\text{B}$ of the depropanizer column shown in Figure~\ref{fig:dep_plant_scheme}.

The studied depropanizer column processes the feed mixture of nine hydrocarbons C3--C5. The purpose of this column is to separate the feed into C3-fraction-rich distillate product $x_\text{D}$ and to C4/C5-fraction-rich bottom product $x_\text{B}$. The available operational degrees of freedom are feed flowrate $F$, bottom product flowrate $B$, distillate flowrate $D$, reflux flowrate $R$, heat duty in the reboiler $Q_\text{B}$, and heat duty in the condenser $Q_\text{D}$. Most of these variables are available as historical data. These are marked correspondingly in Figure~\ref{fig:dep_plant_scheme}. The plant measurements, also available from historical data, are pressure at the top of the column $p_\text{D}$, pressure at the bottom of the column $p_\text{B}$, and temperatures of distillate $T_\text{D}$, of bottoms $T_\text{B}$, at the top of the distillation column $T_\text{C,D}$ and at the bottom of the distillation column $T_\text{C,B}$. The vector of eleven available input variables is given as:
\begin{equation}
	m = \left(F, R, Q_\text{B}, p_\text{D}, p_\text{B}, T_\text{D}, T_{\text B}, T_\text{C,D}, T_\text{C,B}, \frac{R}{F}, \frac{Q_\text{B}}{F}\right)^\intercal.
\end{equation}
The use of the thermodynamic properties model to monitor top/bottom stream compositions is prohibitive in this case, even under any appropriate ideality assumptions.  This is because there are too many degrees of freedom for the treated multi-component mixture that cannot be inferred from plant data. The current inferential sensor (denoted as Ref), applied in the refinery, uses three out of eleven variables and is designed according to~\cite{king_2011} as follows:
\begin{equation} \label{eq:ref_model_FCC}
	x_\text{B} = a_0 + a_1 p_\text{B} + a_2T_\text{C,B} + a_3 \frac{Q_\text{B}}{F},
\end{equation}
where $a_0$ is an intercept, a so-called bias term.


This problem represents a rather standard and well-studied case study of designing an inferential sensor.

\subsection{VGH Unit}
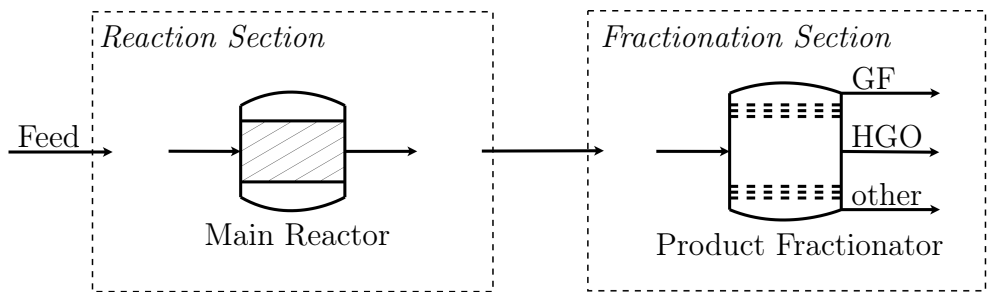
\begin{figure}
	\centering
	\begin{tikzpicture}
		\node[anchor=south west,inner sep=0] (image) at (0,0) {\includegraphics[width=\linewidth]{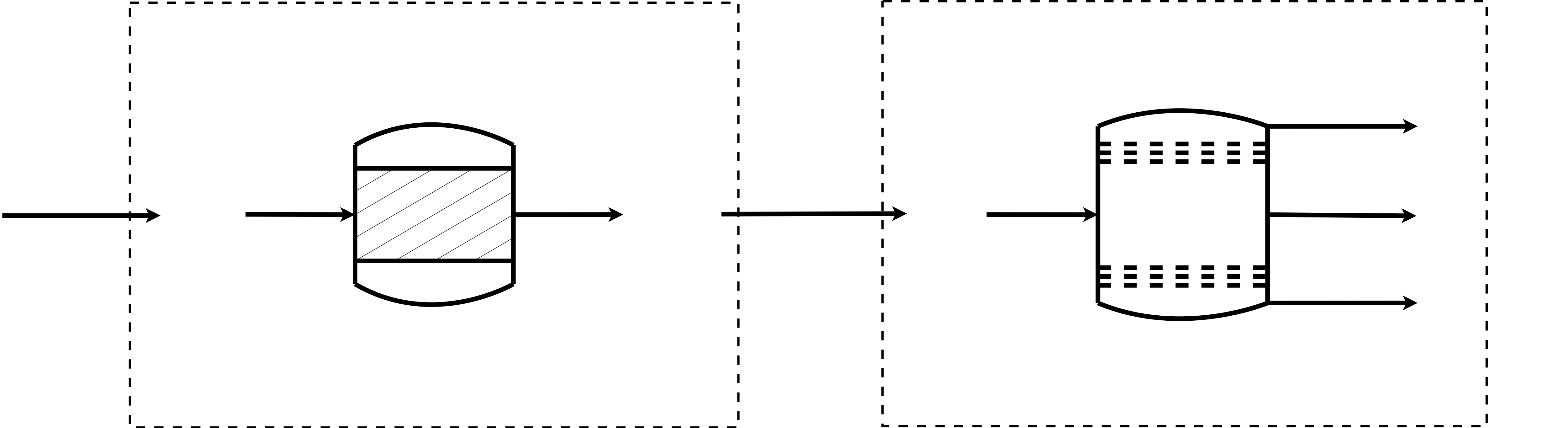}};
		\node at (78pt,97pt) {\emph{Reaction Section}};
		\node at (280pt,97pt) {\emph{Fractionation Section}};
		\node at (16pt,59pt) {Feed};
		\node at (110pt,22pt) {Main Reactor};
		\node at (328pt,81pt) {GF};
		\node at (333pt,59pt) {HGO};
		\node at (333pt,37pt) {other};
		\node at (300pt,18pt) {\parbox{5cm}{\centering Product Fractionator}};
	\end{tikzpicture}
	\caption{A schematic diagram of the VGH unit.}
	\label{fig:VGH_plant_scheme}
\end{figure}

The purpose of this unit is to process the vacuum distillates by hydrotreating. This unit is separated into a high-pressure reaction section and a low-pressure fractionation section (see scheme in Figure~\ref{fig:VGH_plant_scheme}). The main part of the reaction section is represented by the main reactor that hydrogenates the feed. This operation refines the feed from impurities, e.g., nitrogen and sulfur. The reaction section feeds the downstream fractionation section. Here the products are separated into a gasoline fraction (GF), a hydrogenated gasoil (HGO) and other (secondary) products.

Beside the main reactor, the VGH unit involves dozens of low-/high-pressure tanks, heat exchangers, coolers and several distillation columns and furnaces. Furthermore, the unit contains many sensors and control (mostly PI controllers) devices and instrumentation to provide desired operating conditions and products. Overall, there are approximately 1,000 historical values available. Therefore, the inferential sensor design for the VGH unit represents a much more challenging problem compared to the case of the FCC unit (11 variables measured at one distillation column).

The variable to be inferred by the soft sensor is HGO product purity expressed in terms of 95\,\% point of distillation curve $T_{95\%,\text{HGO}}$. The design of an inferential sensor is performed on the subset of the input variables selected from the whole available dataset. The candidate inputs are selected based on consultation with operators and plant management. The resulting set of 30 candidate inputs is following:
\begin{align}\label{eq:full_sensor_vgh}
	m = & \left(PCT_{\text{HGO}}, PCT_{\text{GF}}, T_{\text{ex,1}}, T_{\text{ex,2}}, T_{\text{ex,3}}, T_{\text{ex,4}},\right. \notag\\ 
	& T_{\text{wabt,1}}, T_{\text{wabt,2}}, T_{\text{wabt,3}}, T_{\text{wabt,4}}, T_{\text{wabt,5}}, \notag\\
	& RX_1, RX_2, RX_3, RX_4, RX_5, RX_6, RX_7, \\
	& x_{\text{H}2}, T_{\text{frac,1}}, T_{\text{frac,2}}, F_{\text{frac,heat}}, p_{\text{frac}}, \notag\\
	& \left. F_{\text{f,rec}}, F_{\text{f}}, x_{\text{f,N2}}, x_{\text{f,S}}, T_{\text{f,5p}}, T_{\text{f,50p}}, T_{\text{f,95p}}\right)^\intercal,\notag
\end{align}
with pressure-compensated temperature $PCT$, exotherms for the reactors $T_{\text{ex}}$, weighted average bed temperatures in the reactors $T_{\text{wabt}}$, ratios of gas/liquid phases in different sections $RX$, content of the hydrogen in the reaction section $x_{\text{H}_2}$, temperatures in the main fractionator $T_{\text{frac}}$, flow rate of heat medium for main fractionator $F_{\text{frac,heat}}$, pressure in the main fractionator $p_{\text{frac}}$, feed flowrate reconciled  $F_{\text{f, rec}}$, feed flowrate $F_{\text{feed}}$ and content of impurities in the feed $x_{\text{f,N2}}$, $x_{\text{f,S}}$, $T_{\text{f}}$. 

The pressure-compensated temperature is calculated according to Clausius-Clapeyron equation~\citep{king_2011}:
\begin{equation} \label{eq:PCT}
	PCT = \frac{1}{\frac{R}{H_v}\ln{\frac{P}{P_{\text{ref}}}}+\frac{1}{T}},
\end{equation}
where $R$ is the universal gas constant, $H_v$ is a heat of vaporization, $P$ is an absolute pressure, $P_\text{ref}$ is a reference pressure and $T$ is the absolute temperature.

Current inferential sensor (Ref) used in the refinery is of the following linear structure: 
\begin{equation} \label{eq:ref_model_VGH}
	T_\text{95\%,HGO} = a_0 + a_1 PCT_{\text{HGO}}.
\end{equation}
This seemingly simple inferential sensor is actually a nonlinear soft-sensor. The operators in the refinery have a good past experience with its performance. However, some recent operating conditions and changes to feedstock in the VGH unit caused significant deviations between estimated values from the reference inferential sensor and the values obtained by the lab analysis. The plant management is unsure about the cause and so this study looks at the whole unit and its operation within up- and down-stream sections.

\section{Preliminaries}\label{sec:preliminaries}
This section introduces raw data pre-processing methods, methods for multivariate data treatment, and selected methods of soft sensor design. The analyzed dataset includes $n$ measurement points, therefore:
\begin{equation} \label{eq: generalisation_m_y}
	M:=\begin{pmatrix}
	m_1^\intercal\\ m_2^\intercal\\ \vdots\\ m_n^\intercal
	\end{pmatrix},
	\quad y:=
	\begin{pmatrix}
	y_1\\ y_2\\ \vdots\\ y_n
	\end{pmatrix},
\end{equation}
where $M$ is a matrix of input dataset and $y$ is a vector of output variable measurements.

\subsection{Data Pre-processing}\label{sec:normalization}
The multivariate dataset usually contains data on different scales, e.g., due to standards applied in the company for data units. This can inhibit a proper analysis of the dataset covariance and of the impact of variables in the analyzed system. A step to reduce the discrepancy between the variables is the data pre-processing involving the centering and normalization of the data.

The mean-centered data can be obtained by:
\begin{equation} \label{eq:centering}
	M_\text{C} = M - 1\bar m^\intercal, \quad y_\text{C} = y - \bar{y},
\end{equation}
where $M_\text{C}$ is a matrix of centered input dataset, $1$ is a vector of ones in $\mathbb R^n$, $\bar{m}\in\mathbb R^{n_p}$ is a mean vector of $M$ taken column-wise, $y_\text{C}$ is a vector of centered output variable measurements and $\bar{y}$ is a mean of the output variable measurements. Subsequently, data normalization ($M_\text{C}\to M_\text{N}$, $y_\text{C}\to y_\text{N}$) can be performed such that the data values lie in a desired interval. A commonly used interval is $[-1, 1]$. Another option is the standardization of the data (e.g., required for PCA) to have a zero mean and a unit variance~\citep{kadlec_2009}.

\subsection{Data Treatment Methods}\label{sec:data_treatment}
Industrial data contains systematic and random errors~\citep{su_2009}. The presence of systematic errors in measurements is caused by the non-standard and infrequent situations in the industrial unit, which can be expected (e.g., maintenance) or unexpected (shutdown or plant tripping). 
Another significant source of systematic errors is failures and inaccuracies (measurement bias) of the sensors.

The detection of a certain class of systematic errors can be carried out through visual inspection in time series plots~\citep{alves_2007}. If the same interval of significantly deviated measurements is distinct in all variables, it suggests a potential source of systematic errors and it needs to be omitted before the design of the inferential sensor. Unlike other errors, this situation is easy to detect using a bare eye.

On the other hand, there are situations when one (or some) of the online sensors is suddenly broken or malfunctions. Such systematic errors can be difficult or impossible to indicate by visual inspection in time series plots. This section refers to several multivariate data treatment methods to reduce the number of systematic errors remaining after the visual detection.
 
\subsubsection{Hotelling's $T^2$ Distance}
The Hotelling's $T^2$ distance ($T^2$ distance) is {based on the distribution} developed by H.~Hotelling~\citep{hotelling_1931}. This distribution is a generalization of Student's $t$-distribution. The $T^2$ distance allows us to analyze multivariate datasets and to detect outliers within the set of different variables, e.g., temperatures, pressures, or flow rates. The key aspect of this distance is a covariance among variables involved in the analyzed data expressed by variance-covariance matrix $S$:
\begin{equation}\label{eq:cov_matrix}
	S = \frac{1}{n-1}{M_{\text{N}}} M_{\text{N}}^\intercal.
\end{equation}
The covariance matrix is used to determine Hotelling's $T^2$ distance for each data point:
\begin{equation}\label{eq:T2_dist}
	d_{T^2} = \begin{bmatrix}
		(m_{\text N,1} - \mu)^\intercal S^{-1} (m_{\text N, 1} - \mu) \\
		\vdots \\
		(m_{\text N, n} - \mu)^\intercal S^{-1} (m_{\text N, n} - \mu) \\
	\end{bmatrix},
\end{equation}
where $m_{\text N, i}$ is a vector of measured variables for $i^{\text{th}}$ sample point and $\mu$ is a mean of the sample. 

If the data is normalized (see previous section), $\mu = 0$. According to the values of $d_{T^2}$, it is possible to determine the most deviated measurements (outliers) from the center. The condition to separate admissible and inadmissible measurements by $T^2$ distance is usually set as the empirical $3\sigma$ rule of thumb (probability to include 99.7\,\% measurements) or using $\chi^2$ test but some tuning might be needed based on the data quality.

\subsubsection{Minimum Covariance Determinant}
The Minimum Covariance Determinant (MCD) method~\citep{rousseeuw_1984} is one of the first tools for the outliers detection with high robustness. The distance metric of MCD is the so-called Mahalanobis distance given by the following equation:
\begin{equation}\label{eq:MCD_dist}
	d_{\text{MCD}} = \begin{bmatrix}
		\sqrt{(m_{\text N,1}-\mu)^\intercal S^{-1} (m_{\text N,1}-\mu)} \\
		\vdots \\
		\sqrt{(m_{\text N,n}-\mu)^\intercal S^{-1} (m_{\text N,n}-\mu)} \\
	\end{bmatrix},
\end{equation}
which is closely related to Hotteling's $T^2$ distance~\eqref{eq:T2_dist}. Despite the similarity of the distance metrics of these methods, the principle of MCD is quite different from Hotteling's $T^2$ method. MCD looks for the subset of measurements with the minimum determinant of the corresponding covariance matrix. In other words, the resulting subset of measurements should occupy the smallest volume possible (determinant of the covariance matrix). The algorithm can be viewed as an enhancement of Hotteling's $T^2$ distance method.

The iterative algorithm of MCD starts with a random guess of the initial subset. Subsequently, the mean $\mu$ and covariance matrix $S$ of the initial subset are calculated. According to the calculated $\mu$ and $S$, it is possible to evaluate $d_{\text{MCD}}$ from~\eqref{eq:MCD_dist} for each measurement (not only for the selected subset). Subsequently, the new subset of $h$ measurements with the smallest distances $d_{\text{MCD}}$ is selected from the whole set. If the covariance determinant of the new subset is decreased compared to the covariance determinant of the previous subset, the new subset is used in the next iteration of the MCD algorithm. Otherwise, the sought subset has been found (as the previously selected subset) and the MCD algorithm is terminated. The tuning parameter of this scheme is represented by the least number of the retained measurements $h$ from the treated dataset. This parameter is usually adjusted according to the interval $\frac{n + n_\text{p} + 1}{2} \leq h \leq n$~\citep{hubert_2010} or it can be adjusted by the user e.g., based on the visual inspection of the time series of some crucial variables.

Due to the random character of this method, it is desired to perform several runs with different initial guesses to avoid local minima. According to the results from different runs of the MCD method, it is possible to derive a final subset. The vector of distances $d_{\text{MCD}}$ is evaluated in each iteration. The measurements with the smallest distances create a new subset for the next iteration of MCD. This process is terminated when the determinant of the covariance matrix does not decrease anymore.

The mean $\mu$ and the covariance matrix $S$ of the final subset are subsequently used to evaluate $d_\text{MCD}$ from~\eqref{eq:MCD_dist} for the whole set. According to the values of $d_\text{MCD}$, it is possible to determine the most deviated measurements (outliers) from the center. The condition to separate admissible and inadmissible measurements by MCD is established by appropriate distribution~\citep{hardin_2005} considering the desired confidence level.

\subsubsection{$k$-means Clustering}
The $k$-means clustering method~\citep{forgy_1965} separates measurements from the multivariate dataset into different groups (clusters). Each measurement is assigned to the cluster according to the closest center of a cluster (i.e., mean of the cluster data points). The area of clusters should be as small as possible yet the data points in the different clusters should be as far from each other as possible. 
 
The $k$-means clustering is able to adjust the performance by using different distance metrics $d_{\text{CL}}$. The frequently used distance metric is the squared Euclidean distance in the form:
\begin{equation}\label{eq:kmeans_dist}
		d_{\text{CL},i,j} = (m_{\text N, i,j}-\mu_j)^\intercal(m_{\text N,i,j}-\mu_j),
\end{equation}
where $\mu_j,\ j\in\{1, 2, \dots,  k\}$ is the center of the $j^{\text{th}}$ cluster representing the mean of the corresponding data points and the index $i$ characterizes the ordinal number of the data points within the $j^{\text{th}}$ cluster.

The selection of the desired number of clusters ($k$) is highly related to the data quality. In trivial cases, it is possible to determine the value of $k$ by visual inspection, where the data points create visible groups representing different operating conditions of the unit. In a non-trivial case, it is possible to determine the value of $k$ according to the elbow method or using various goodness of fit criteria~\citep{kodinariya_2013}.

The algorithm of $k$-means clustering is initiated by a random guess of the locations of the desired centers. Therefore several runs of the algorithm should be performed with different initial guesses. The measurements should be assigned to the clusters with the highest frequency of the assignment from the different runs of the algorithm, similar to the final subset of the MCD method. Once the final clusters are created, it is possible to determine the outliers as measurements in the particular cluster. It can be seen that the clusters predominantly constituted by outliers contain a smaller amount of data compared to the rest of the clusters.

According to the nature of the aforementioned data treatment methods, one could expect the performance of MCD at least as good as the performance of the $T^2$ distance method. The $k$-means clustering can outperform the rest of the methods if measurements involve several clearly distinct operating points (steady states) of a particular unit.

\subsection{Methods for Inferential-Sensor Design}\label{sec:design_IS_methods}
We study data-driven methods for soft-sensor design. Each method solves two sub-problems: the structure selection of the soft sensor and the calculation of soft sensor parameters. The investigated methods are based on different principles: on the analysis of the variance-covariance matrix of the dataset (PCA and PLS) and on sparsity enforcement (LASSO and subset selection).

An effective design procedure usually requires splitting the  available dataset (input matrix $M_\text{N}$, output vector $y_\text{N}$) into the following subsets: dataset for sensor design that contains training data $\left(M_\text{N}(\mathcal{T}), y_\text{N}(\mathcal{T})\right)$ and dataset used for the performance evaluation of designed soft sensors that contains testing data $\left(M_\text{N}(\mathcal{S}), y_\text{N}(\mathcal{S})\right)$. Here $\mathcal{T}$ and $ \mathcal{S}$ denote the corresponding row-selection operators.

\subsubsection{Ordinary Least Squares Regression}
The basic method of soft-sensor design is Ordinary Least-Squares Regression (OLSR). This method estimates the parameters of an inferential sensor according to
\begin{equation}\label{eq:olsr}
	\min_a \frac{1}{2}\sum_{\forall i\in\mathcal T}(y_{\text N, i}-m_{\text N, i}^\intercal a)^2 \equiv \min_a \frac{1}{2}\|y_\text N(\mathcal T)-M_\text N(\mathcal T)a\|_2^2,
\end{equation}
which minimizes the sum of squared errors between measurements and sensor predictions.

The method can potentially result in a sparse sensor structure, e.g., when strong linear dependencies exist among some variables. One can thus talk about OLSR being able to select (sparsify) the sensor structure. Generally, OLSR cannot effectively (or actively) strive against overfitting. Its performance has to be usually enhanced in combination with other methods that consider not only the accuracy but also the complexity of the resulting model.

\subsubsection{Principal Component Analysis}
Principal Component Analysis (PCA)~\citep{pearson_1901} is a method of identifying an $\tilde n_\text{p}$-dimensional subspace ($\tilde n_\text{p}\leq n_\text{p}$) of orthogonal coordinates that exhibit a maximum variance in a given dataset. 

The principal components are identified by the eigendecomposition of the covariance matrix $S$ of the mean-centered, unit-variance data by taking the eigenvectors (for subset definition) and the associated eigenvalues (for measure of variance). Each eigenvector represents one principal component that explains a certain amount of the data variance. The desired amount of total variance can be captured by selecting several ($\tilde n_\text{p}$) principal components within the eigenvectors with maximum explained variance. The regression is then carried out over the selected (principal components) subspace using~\eqref{eq:olsr} with $\tilde n_\text{p}$ parameters.

The usage of PCA regression represents an advantage mainly in the case of an insufficient amount of the output data. In fact, this is the usual situation in the industry, where the measurement of the desired output variable is too expensive or rare. Such situation leads to performance deterioration of many data-driven methods for soft-sensor design as they usually require large number of measurements. The PCA regression method has gained its popularity because of being able to learn from the measurements of the online sensors and thus being able to outperform other data-driven methods in certain cases.

\subsubsection{Partial Least Squares}
Partial Least Squares (PLS) regression~\citep{wold_1984} is a statistical method searching for a linear regression model of predicting the output (predicted) variable using input variables by the projection into a new space of principal components. Although PLS regression is not an unsupervised learning approach (as PCA), the principle of these methods is essentially the same~\citep{wold_1984} and both methods are intended to reduce the dimensionality of the problem. 

The most common approaches for PLS regression are nonlinear iterative partial least squares (NIPALS) and SIMPLS~\citep{dejong_1993}. Both approaches iteratively calculate the principal components. The principal components are calculated by SVD decomposition of the following cross-covariance matrix $S_{\text{cross}}$:
\begin{equation}\label{eq:cross_cov_matrix}
	S_{\text{cross}} = \frac{1}{n-1}{M_{\text{N}}} y_{\text{N}}^\intercal.
\end{equation}
The selection of the desired number of principal components is then performed in the same way as in the case of PCA. Subsequently, the principal components are used to design the soft sensor much like in the case of PCA.

\subsubsection{Least Absolute Shrinkage and Selection Operator}
Least Absolute Shrinkage and Selection Operator (LASSO)~\citep{santosa_1986} is a method that simultaneously identifies the structure of the model and its parameters by solving the following optimization problem:
\begin{equation}\label{eq:lasso}
	\min_a \ \frac{1}{2}\sum_{\forall i\in\mathcal T}(y_{\text N, i}-m_{\text N, i}^\intercal a)^2 + \lambda\|a\|_1,
\end{equation}
where $\lambda$ is a weight between the accuracy of the model training and the model overfitting. The magnitude of the $\ell_1$-penalization element results in certain parameters being equal to zero at the optimum of~\eqref{eq:lasso}. The resulting model is then less complicated and usually more interpretable.

The LASSO technique belongs to the regularised regressions family. Beside the LASSO regression, this family involves many other methods, but the most important ones are ridge regression~\citep{hoerl_1970} and elastic net~\citep{zou_2005}. The ridge regression has a similar objective function compared to LASSO. However, the ridge regression uses the $\ell_2$-penalization element to reduce the value of all parameters. The usage of ridge regression is preferred when the input variables are highly correlated. The elastic net technique effectively combines LASSO and ridge regression. It weighs between $\ell_1$-penalization element and $\ell_2$-penalization element within the objective function.

\subsubsection{Optimal Subset Selection with Model-overfitting Criteria}
Subset selection denotes a class of methods that explicitly seek for the simplest possible sensor structure such that some model-overfitting criterion $J(a, z)$ is minimized~\citep{miyashiro_2015}. Here the variable $z$ denotes a vector with binary entries $z\in\{0, 1\}^{n_\text{p}}$ signifying the selection of $j^\text{th}$ input into the sensor structure. Correspondingly, the sum of the vector entries $\sum_{j=1}^{n_\text{p}}z_j=1^\intercal z$
denotes the sensor complexity.

Optimal subset selection solves the following bi-level program~\citep{ber_2016}:
\begin{subequations}\label{eq:SS_overfit}
	\begin{align}
		& \min_{a,\, z\in \{0, 1\}^{n_\text{p}}} \ J(a, z) \\
		& \ \text{s.t. } a \in \text{arg}\min_{\tilde a} \frac{1}{2}\|y_\text N(\mathcal T) - M_\text N(\mathcal T)\tilde a\|_2^2 \\
		& \quad \ -\bar a z_j \leq \tilde a_j \leq \bar a z_j, \forall j\in\{1,\dots,n_\text{p}\},
	\end{align}
\end{subequations}
where $\bar a$ represents an upper bound on $\|a\|_\infty$ to be tuned and the optimization criterion $J(\cdot)$ might take the form ($\text{RSS}:=\|y_\text N(\mathcal T) - M_\text N(\mathcal T)a\|_2^2$):
\begin{equation}\label{eq:r2adj}
J_{R^2_{\text{adj}}} = \frac{\text{RSS}}{n - 1^\intercal z - 1}, 
\end{equation}
or
\begin{equation}\label{eq:aicc}
J_{\text{AIC}_\text{C}} = n\log\frac{\text{RSS}}{n} + 2(1^\intercal z),
\end{equation}
or
\begin{equation}\label{eq:bic}
 J_{\text{BIC}} = n\log\frac{\text{RSS}}{n} + \log(n)(1^\intercal z). 
\end{equation}
The bi-level program (Eq.~\eqref{eq:SS_overfit}) can be effectively resolved by standard MIQP solvers using big-M reformulation as shown in~\cite{takano_2020}.

\subsubsection{Optimal Subset Selection with Cross-Validation Criterion}
The principle of Subset Selection with Cross-Validation Criterion (SS-CV) is to mimic a standard cross-validation procedure within the training dataset. Let us divide the training data into $K$ smaller subsets $\mathcal{N}_k$, such that:
\begin{align}\label{eq:SS_cross_subsets}
	\mathcal{T} = \bigcup_{k \in K} \mathcal{N}_k, \quad \mathcal{N}_k \cap \mathcal{N}_{k'} = \emptyset, \ \ \forall k \neq k', \quad K \geq 2.
\end{align}
The data is distributed into training ($\mathcal{T}_k$) and validation ($\mathcal{V}_k$) sets as follows:
\begin{align}\label{eq:SS_cross_training_validation_japanese}
	\mathcal{V}_k:=\mathcal{N}_{k}, \quad \mathcal{T}_k := \mathcal{T} \setminus \mathcal{N}_k, \quad \text{card}(\mathcal{T}_k) \geq n_\text{p}, \ \ \forall k \in K.
\end{align}
where $\mathcal{V}_k$ sets contain unique data, while the different $\mathcal{T}_k$ sets involve recurring measurements.
The optimal SS with cross-validation solves~\citep{takano_2020}:
\begin{subequations}\label{eq:SS_cross}
	\begin{align}
		&\min_{a^{(k)}, \forall k\in K,\ z\in \{0, 1\}^{n_\text{p}}} \ \frac{1}{2}\sum_{k=1}^{K}\|y_\text N(\mathcal{V}_k) - M_\text N(\mathcal{V}_k)a^{(k)}\|_2^2 \\
		& \qquad \text{s.t. } \forall k\in K\!:\ a^{(k)} \in \text{arg}\min_{\tilde a} \frac{1}{2}\|y_\text N(\mathcal{T}_k) - M_\text N(\mathcal{T}_k)\tilde a\|_2^2\\   & \qquad \qquad \qquad \qquad \qquad \text{s.t. } -\bar az_j \leq \tilde a \leq \bar az_j, \ \forall j\in\{1,\dots,n_\text{p}\}.
	\end{align}
\end{subequations}

The problem~\eqref{eq:SS_cross} can be solved for several values of $K$---considering constraints on parameter identifiability, i.e., the cardinality condition in Eq.~\eqref{eq:SS_cross_training_validation_japanese}---and for different randomly generated distributions of data into $\mathcal{T}_k$ and $\mathcal{V}_k$ sets. The structure of the resulting sensor is then given by the most frequent inputs occurring in the calculated sensors. Once the optimal sensor structure is calculated, a least-squares fitting of such a model is used with the entire training dataset to determine the parameters of the designed soft sensor. Similarly to problem~\eqref{eq:SS_overfit}, the problem~\eqref{eq:SS_cross} can be effectively resolved by standard MIQP solvers.

\section{Results}\label{sec:results}
We present the results for both the presented use cases. We compare the performance of the presented data treatment methods and methods for soft-sensor design. Due to data confidentiality, the graphical representations of the results use the normalization of variables in the interval $[0, 1]$.

\subsection{Implementation details}
The implementation of all the presented methods is performed in MATLAB. For the initial data treatment, we use the Hotelling's $T^2$ distance considering $\chi^2$-distribution with the probability of including 99.7\,\% measurements. For the MCD method, we select the value of parameter $h$ as a midpoint of the interval $\frac{n + n_\text{p} + 1}{2}\leq h \leq n$~[1]. The outliers are determined by MCD considering an approximation of $F$-distribution~[2] with the same probability as in the $T^2$ distance method. As a preliminary analysis suggested, the industrial data seem to be not normally distributed. Therefore the $T^2$ distance method considering $\chi^2$-distribution tends to remove larger portions of data than MCD with $F$-distribution. The number of the desired clusters for the $k$-means clustering is determined using the elbow method. The results of the MCD method and the $k$-means clustering are gathered and averaged over 100 different runs of the respective algorithms. This is because of the inherent randomness of these methods, as mentioned above.

For the soft-sensor design, we set the variance-covariance methods (PCA and PLS) to select the amount of the variance explained by the principal components to at least 98\,\%. The PLS method uses SIMPLS approach from MATLAB. We use Yalmip~\citep{yalmip} and Gurobi~\citep{gurobi} to solve various instances of the problems~\eqref{eq:olsr}, \eqref{eq:lasso}, \eqref{eq:SS_overfit}, and~\eqref{eq:SS_cross}. 

We will study two different scenarios of soft-sensor design for each use case \wrt splitting the data into training and testing subsets. In both scenarios, the training set is used for the design of the reference as well as the rest of the studied inferential sensors. In the first scenario, the available dataset is divided into subsets based on the time series. The data from an earlier time period is used for training and the data from a later time period is used for testing. This situation simulates soft-sensor design at a certain point in time using historical (training) data. The testing phase then mimics the future sensor performance, where the sensor is employed without any adaptation of its structure despite possible variations in plant operating conditions.

The second scenario groups the available data among training/testing subsets randomly. The results thus reveal the potential of the studied sensor-design methods for adaptation of the sensor structure to the changing operating conditions. In this scenario, the final results are gathered from 50 runs with different training/testing dataset distributions.

In order to tune the value of $\lambda$ in~\eqref{eq:lasso} we use the goodness-of-fit criteria~\eqref{eq:r2adj}~--~\eqref{eq:bic} and cross-validation on the training set. We first obtain the candidate values of $\lambda$ that minimize one of the goodness-of-fit criteria by training the sensors on the whole training set. Subsequently, we generate twenty different distributions of the training data into two subsets (similar to the SS-CV method). The candidate values of $\lambda$ are used for regression and cross-validation on the generated subsets and the best performing value is used for the final sensor training.

When determining the final design of the soft sensor according to SS with cross-validation, we take a median of $1^\intercal z$ from the results of the different runs (different validation data distribution and different values of $K$: $K \leq 6$ for the FCC unit, $K \leq 4$ for the VGH unit) to obtain the $n_\text{p}^*\leq n_\text{p}$, i.e., the number of inputs of the final sensor. Subsequently, we select the $n_\text{p}^*$ most frequent inputs from the results of the different runs to finalize the sensor structure.

The complexity of each designed soft-sensor structure is determined according to the number of input variables $n_\text{p}^*$. We measure the impact of a particular input on the soft-sensor performance by the value of $\vert a_i \vert$. If the impact of a particular term is less than 0.1\,\% of the maximum value of the desired inferred variable, we neglect the corresponding part of the soft sensor.

The accuracy of the soft sensors is evaluated and compared by the root mean square error (RMSE) of the sensor prediction on the testing dataset. The performance of the industrial soft sensors can be adjusted during the operation by an adaptive bias correction, also called bias update. Many industrial software solutions offer this form of soft-sensor maintenance against the change of operating conditions. The purpose of the bias correction is to improve the accuracy of the succeeding predictions of the soft sensor by adjusting the constant (bias) term $a_0$. The bias is updated when a measurement from a lab analysis is available and when it differs significantly from the sensor prediction~\citep{quelhas_2009}. The frequency of bias correction is thus, on the one hand, a measure of frequency of the change in the plant operating conditions. On the other hand, it reflects the ability of the sensor itself to react to the changes in the operating conditions. The plant operators trust more a soft sensor with less frequent bias updates. Therefore, in addition to the soft-sensor complexity ($n_\text{p}^*$) and accuracy (RMSE), we evaluate the effort of the bias correction (BC) by simulating a bias correction procedure in parallel, i.e., without affecting the prediction error of the sensor evaluated by RMSE. The measure of the bias-correction effort is expressed as the percentage of measurement-based sensor corrections occurrences in the testing dataset.

\subsection{Inferential Sensors for the FCC Unit}\label{sec:res_fcc}
The available historical data involving 32,061 measurement points from online sensors (candidate input variables) represents more than two years of production in the period 2016--2019. This time span contains 181 lab measurements of the bottom product concentration $x_\text{B}$ (output variable). 

\begin{figure}
	\centering
	\captionsetup[subfigure]{labelfont=bf}
	\begin{subfigure}[b]{0.495\textwidth}
		\centering
		\includegraphics[width=\textwidth]{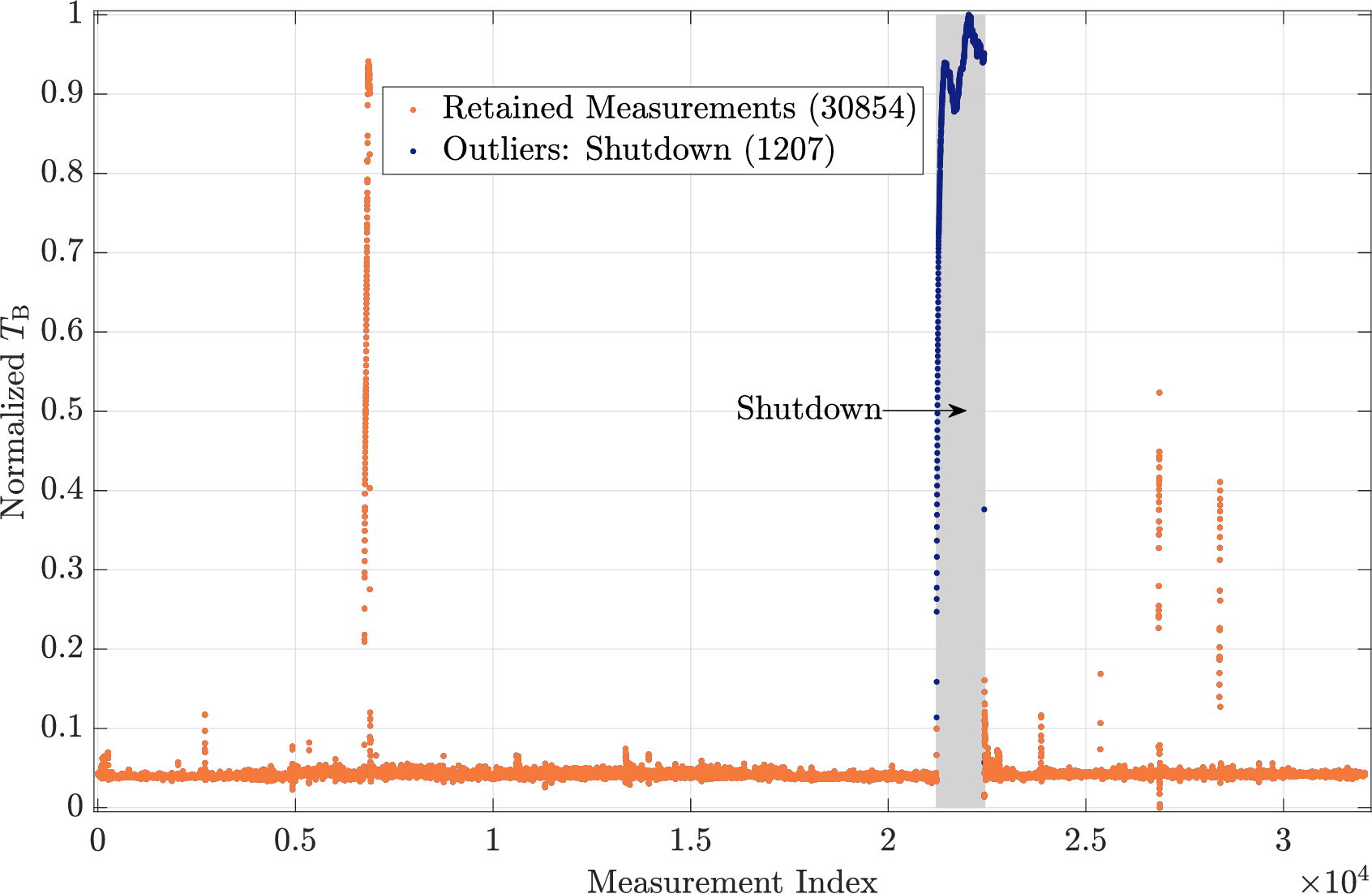}
		\caption{Data pre-treatment by visual inspection detecting plant shutdowns.}
		\label{fig:fcc_GED_shutdown}
	\end{subfigure}
	\hfill
	\begin{subfigure}[b]{0.495\textwidth}  
		\centering 
		\includegraphics[width=\textwidth]{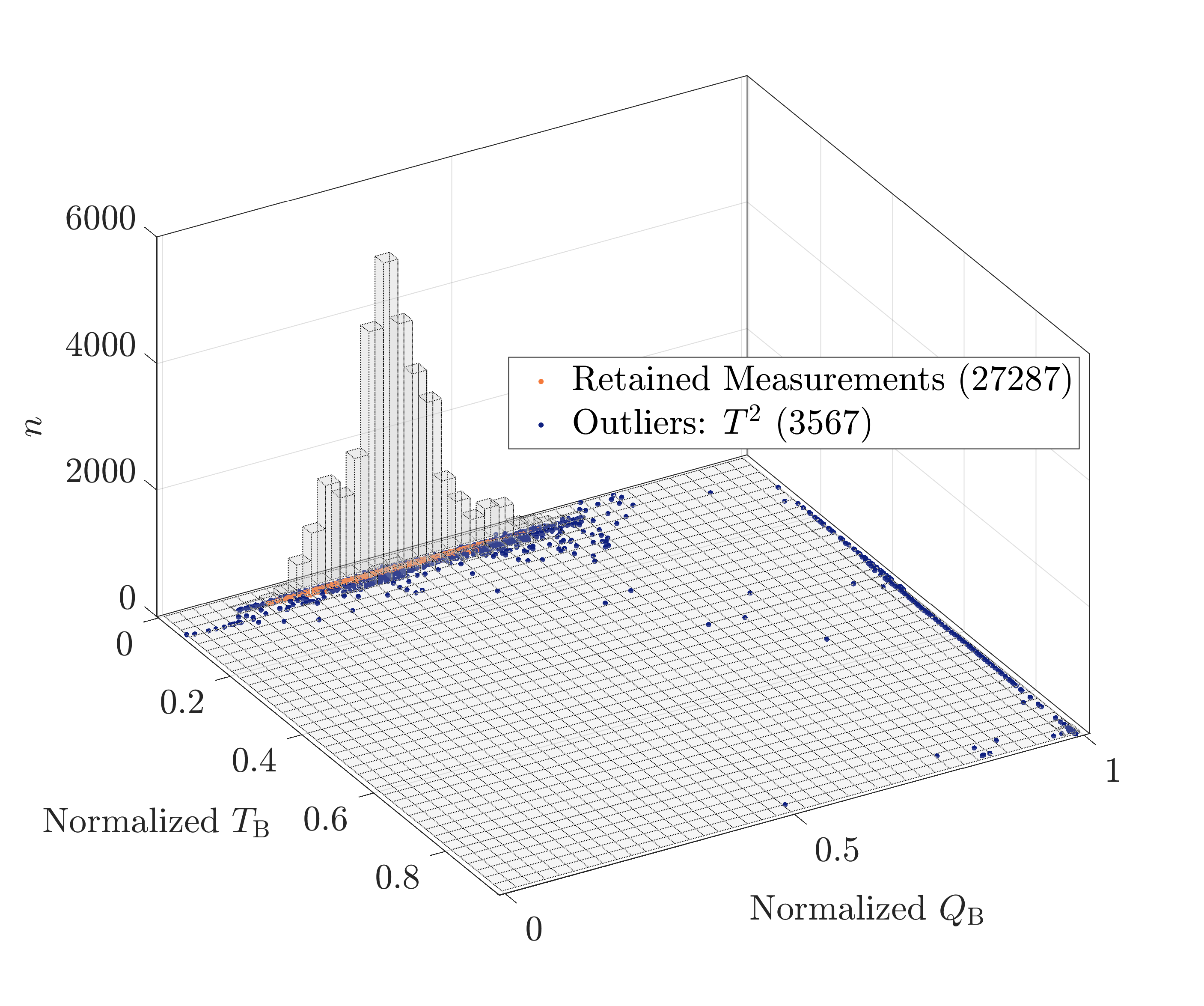}
		\caption{Data treatment by the $T^2$ distance method.}    
		\label{fig:fcc_GED_T2_hist}
	\end{subfigure}
	\vskip\baselineskip
	\begin{subfigure}[b]{0.495\textwidth}   
		\centering
		\includegraphics[width=\textwidth]{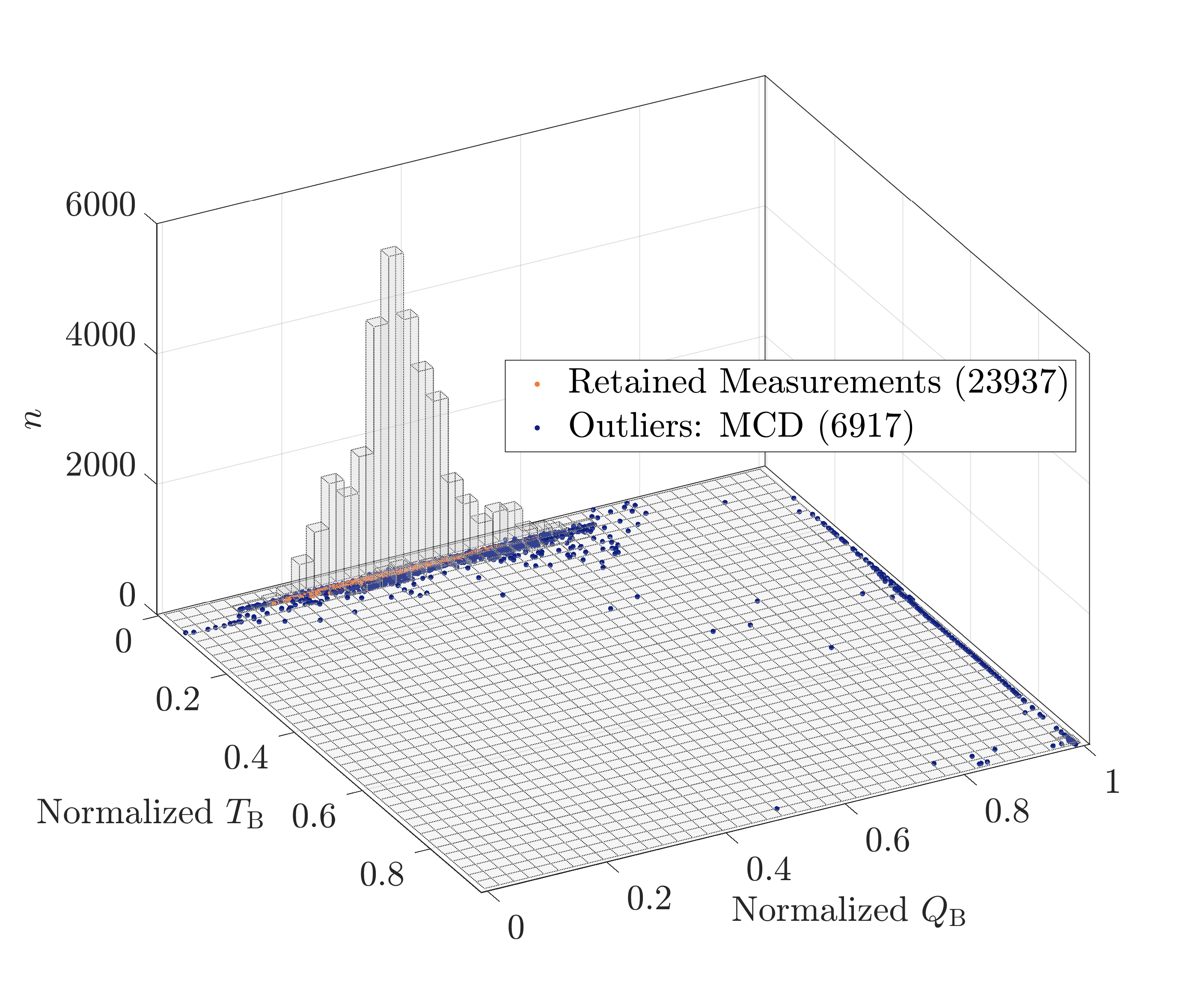}
		\caption{Data treatment by the MCD method.}    
		\label{fig:fcc_GED_MCD_hist}
	\end{subfigure}
	\hfill
	\begin{subfigure}[b]{0.495\textwidth}   
		\centering
		\includegraphics[width=\textwidth]{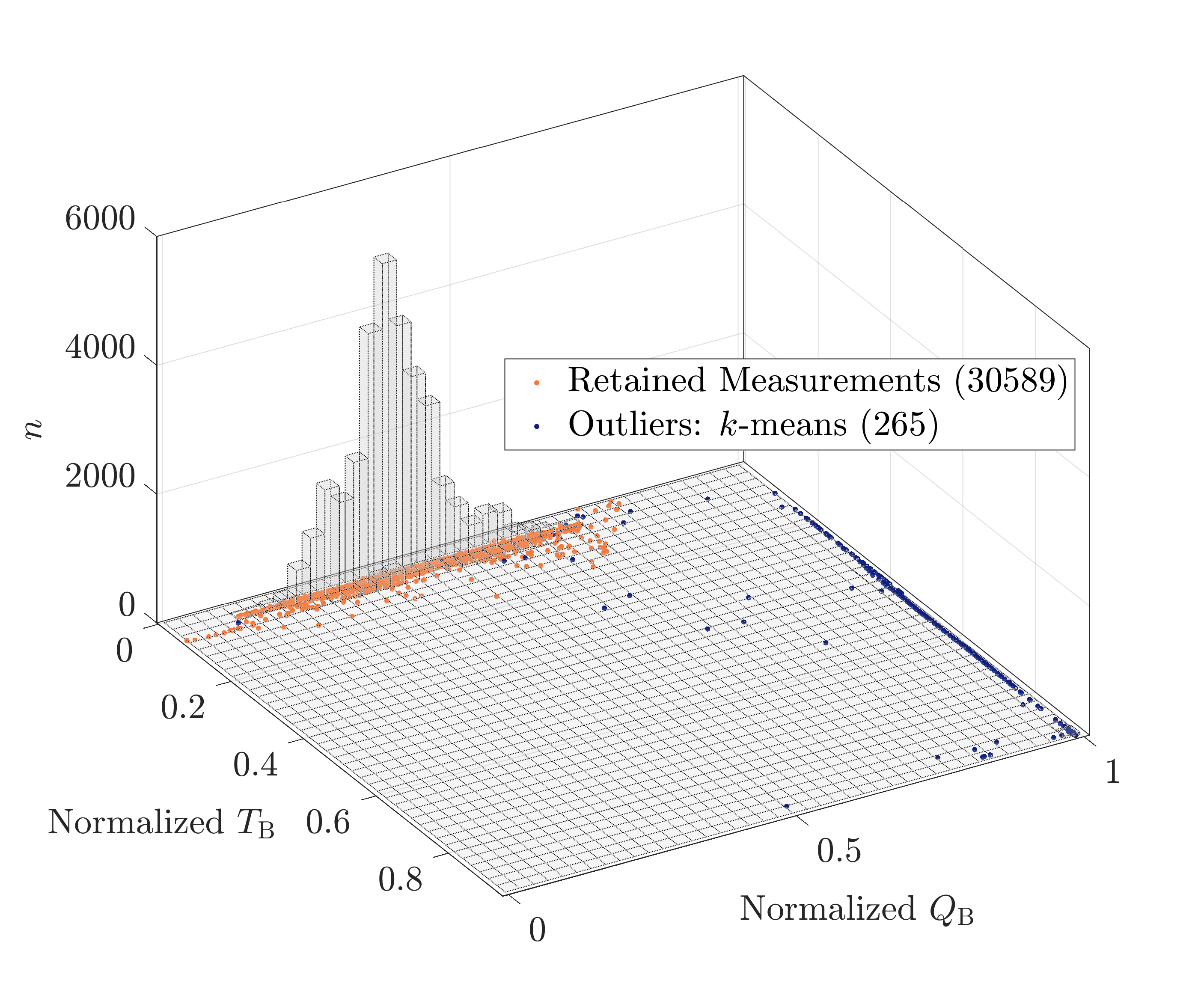}
		\caption{Data treatment by the $k$-means clustering method.}
		\label{fig:fcc_GED_kms_hist}
	\end{subfigure}
	\caption{\textbf{(a)} Normalized bottom product temperature of the FCC unit vs.~measurement index. \textbf{(b)}, \textbf{(c)}, \textbf{(d)} Histogram of the bottom product temperature vs.~reboiler heat duty of the FCC unit and retained measurement vs.~outliers as detected by data treatment methods.}
	\label{fig:fcc_GED}
\end{figure}

We first perform the data treatment to reduce the amount of systematic and gross errors. Figure~\ref{fig:fcc_GED_shutdown} shows visualization of the data treatment results on the normalized temperature of the bottom product $T_\text{B}$. The visual inspection of the time series of the available data (data pre-treatment) reveals the initial set of systematic errors with significantly deviated data, which corresponds to the shutdown period of the unit. This is marked as a thick gray bar in Figure~\ref{fig:fcc_GED_shutdown}. The unit operators confirmed in consultation the correctness of omission of the corresponding 1,207 data points from the further processing.

Subsequently, we applied the $T^2$ distance, MCD, and $k$-means clustering methods to detect outliers in the dataset. The performance of these methods is individually visualized and compared in Figures~\ref{fig:fcc_GED_T2_hist},~\ref{fig:fcc_GED_MCD_hist} and~\ref{fig:fcc_GED_kms_hist} for lucidity. Each figure shows a histogram of data points of bottom product temperature vs.~reboiler heat duty. All the methods clearly identify the most distinct outliers. The results further show that $k$-means clustering (Figure~\ref{fig:fcc_GED_kms_hist}) might be overly conservative as it selected significantly fewer outliers than the other two methods. The low performance of this method is caused by the complex tuning (e.g., number of clusters). The $k$-means clustering method detects only five data clusters, which results in the low number of indicated outliers by this method. The number of outliers indicated by the MCD method (Figure~\ref{fig:fcc_GED_MCD_hist}) is almost twice higher compared to the $T^2$ distance method (Figure~\ref{fig:fcc_GED_T2_hist}). The MCD method thus appears as a reasonable choice here as it removes a significant amount of outliers, yet retains reasonable number of data points, of which it guarantees better quality than the $T^2$ distance approach.

\begin{figure}
	\centering\vspace{.5cm}
	\includegraphics[width=\linewidth]{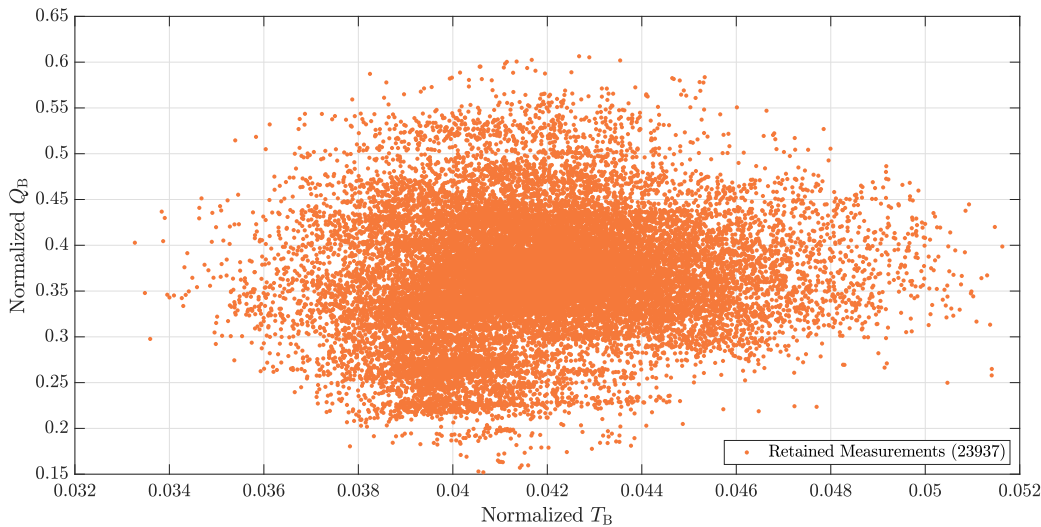}
	\caption{The retained online measurements (by the MCD method) of the bottom product temperature and reboiler heat duty of the FCC unit.}
	\label{fig:fcc_GED_vars_without_GE}
\end{figure} 

It is obvious that the majority of identified outliers (blue points in Figure~\ref{fig:fcc_GED_MCD_hist}) by the MCD method deviates from the area with the highest density of the online measurements. On the same line, the approved measurements (green points in Figure~\ref{fig:fcc_GED_MCD_hist}) are located inside or are very close to this area. This also indicates the good performance of the MCD method. The final set of the retained measurements by this method for the soft-sensor design is shown in Figure~\ref{fig:fcc_GED_vars_without_GE}. It is evident that the MCD method provides well-poised data set, which appears to be close to normal distribution. We can conclude that the available industrial data are of good quality and that the conducted data treatment was able to reveal the high-quality data. 

\subsection{Design of Inferential Sensors for the FCC Unit using Time Series Data}\label{sec:res_fcc_chron}
We first study a scenario where the (chronologically) first 50\,\% of the available data is assigned to the training set and the last 50\,\% of data is assigned to the testing set.

Soft-sensors designed by PCA and PLS require six and seven principal components, respectively, to explain 98\,\% of the variance in the data. This relatively high number of principal components suggests, on the one hand, to use a more complex structure of soft sensor than the reference soft sensor. On the other hand, sensors designed by these methods might be overfitted.

When designing a soft sensor by the SS methods, we compared the performance of the presented overfitting criteria ($\mathit{R}^2_{\text{adj}}$, AIC$_\text{C}$, BIC). We used the principle of parsimony. The simplest sensor yet the best performing one is designed by SS with $\mathit{R}^2_{adj}$ criterion. This sensor is the same as suggested by SS with cross-validation in this case and it is selected for further performance analysis.

\begin{table}
	\caption{Comparison of the number of inputs $n_\text{p}^*$ (number of principal components for PCA and PLS shown in brackets), sensor accuracy (RMSE) and bias correction relative frequency (BC) using time series data for the FCC unit.}
	\label{tab:fcc_res_chron}
	\centering
	\begin{tabular}{c|c c c c c c|c}
		\toprule
		         & OLSR  & PCA     & PLS     & LASSO & SS    & SS-CV & Ref   \\ \midrule
		$n_\text{p}^*$  & 11    & 11\,(6) & 11\,(7) & 5     & 4     & 4     & 3     \\
		RMSE     & 0.120  & 0.096   & 0.104   & 0.099 & 0.099 & 0.099 & 0.117 \\
		BC [\%]  & 29.7 & 21.6   & 24.3   & 20.3 & 23.0 & 23.0 & 28.4 \\
		\bottomrule
	\end{tabular}
\end{table}

A comparison of the designed sensors in terms of their complexity ($n_\text{p}^*$), accuracy (RMSE), and the effort of the bias correction (BC) is shown in Table~\ref{tab:fcc_res_chron}. The results clearly suggest to enrich the structure of reference soft sensor to include at least one extra variable in order to improve its performance (see $n_\text{p}^*$ in Table~\ref{tab:fcc_res_chron}). The least complex sensors are suggested by the LASSO and SS methods. These methods suggest replacing bottom pressure $p_\text{B}$ by temperatures $T_\text{C,D}$ and $T_\text{C,B}$ (LASSO selected also the ratio $R/F$). These sensors (including PCA) exhibit a reduced amount of bias correction compared to all others.

Overall, the accuracy of the reference soft sensor (see RMSE in Table~\ref{tab:fcc_res_chron}) shows almost the worst performance. Only the (most likely overfitted) soft sensor designed by OLSR is worse in this comparison, despite using all the possible eleven inputs. The overfitting by OLSR can be documented by worsened accuracy and also by a high effort of the bias correction.

The highest sensor accuracy is achieved for the PCA-based soft sensor. The improvement compared to the reference soft sensor is approximately 18\,\%. Other proposed advanced sensors show similar performance (improvements of at least 15\,\%). 

Looking at the amount of bias correction, we can see that the most frequently corrected soft sensor is designed by OLSR, while the soft sensor designed by LASSO requires the bias correction less frequently than others. The best sensor would be selected as a compromise between accuracy, complexity, and maintenance (BC) effort. In this respect, all the advanced designed soft sensors represent good candidates.

\begin{figure}
	\centering
	\begin{subfigure}{\textwidth}
		\includegraphics[width=\linewidth]{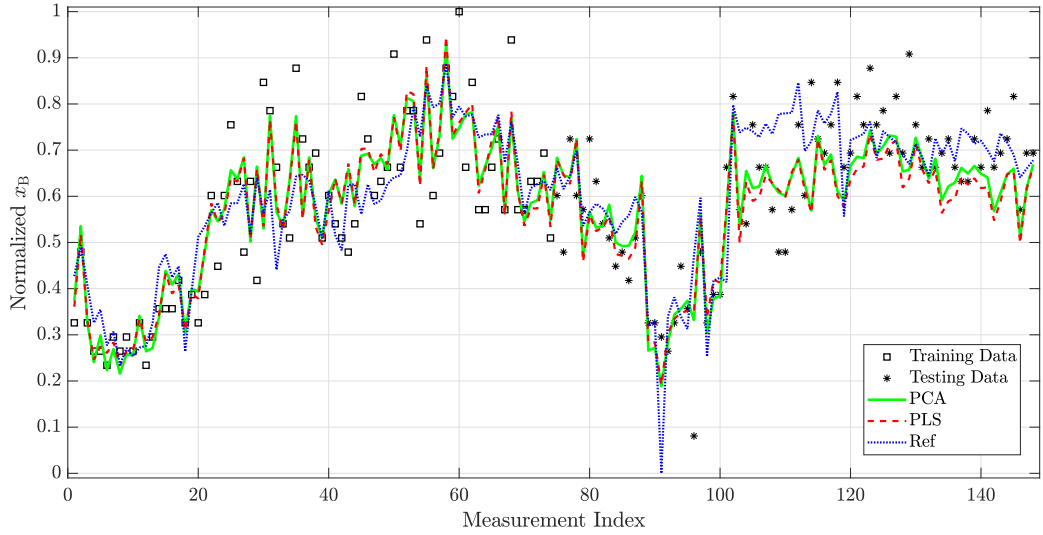}
		\caption{Training and prediction performance of the sensors designed by PCA and PLS methods and of the reference sensor.}
		\label{fig:fcc_chron_IS_PCA_PLS}
	\end{subfigure}
	\begin{subfigure}{\textwidth}
		\vspace{.5cm}
		\includegraphics[width=\linewidth]{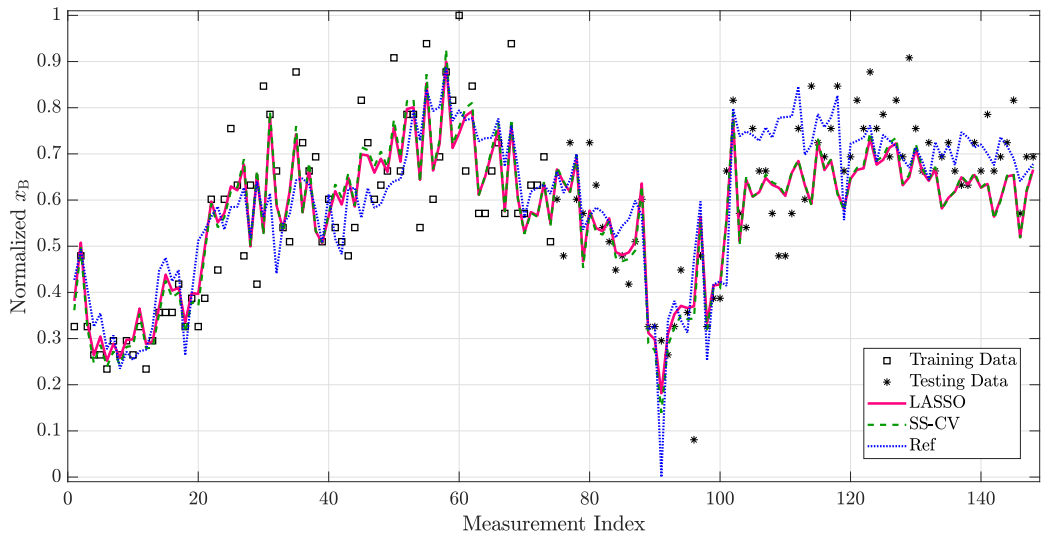}
		\caption{Training and prediction performance of the sensors designed by LASSO and SS-CV methods and of the reference sensor.}
		\label{fig:fcc_chron_IS_LASSO_SS}
	\end{subfigure}
	\caption{Comparison of the soft sensors for the FCC unit designed using time series data.}
	\label{fig:fcc_chron_IS}
	\vspace{0cm}
\end{figure}

In order to provide a more comprehensive comparison of the soft sensors, Figure~\ref{fig:fcc_chron_IS} visualizes their predictive performance on the output variable. The lab-analysis data is shown as black squares (training dataset) and black stars (testing dataset), respectively. The data show significant variability indicating several changes of the operating conditions within the studied time window, in both training and testing datasets. This means that the trained sensors face a rich portfolio of situations and thus a trained sensor can be expectedly valid for a long time after its commissioning. This is confirmed by the aforementioned good performance of the designed sensors and by the relatively low effort of the bias-update mechanism.

Figure~\ref{fig:fcc_chron_IS} further presents the training and testing (predictions) performance of the designed advanced soft sensors, by PCA and PLS (Figure~\ref{fig:fcc_chron_IS_PCA_PLS}; green solid line and red dashed line, respectively) and by LASSO and SS-CV (Figure~\ref{fig:fcc_chron_IS_LASSO_SS}; magenta solid line and green dashed line, respectively), compared in both figures to the reference soft sensor (blue dotted line).

When looking at the performance of the reference sensor in both plots, one can clearly identify several points, where the reference sensor is not able to explain the measurements yet the advanced sensors are. This is present throughout the whole studied time window but it is most evident in the testing phase (around the measurements 80--120).

We can see that despite the behavior of the soft sensors designed by PCA and by PLS being similar in the training phase, the evolution of the predictions of these sensors on the testing data is quite different. This also explains differences in the accuracy and frequency of the bias correction. It also further supports our earlier conjecture of possible overfitting present in these sensors. This observation is in contrast with the bottom plot (LASSO and SS-CV), where the outputs of the visualized advanced sensors are almost identical.

A noticeable part of the testing phase is the last period (around the measurements 130--148), where it seems that the operating conditions in the FCC unit change considerably. There exist corresponding significant discrepancies between the measurements and values inferred by all the advanced soft sensors. The reference soft sensor, however, performs well here, which suggests good robustness properties of this sensor. All the advanced sensors exhibit a slower or faster drift from the measurements. This situation calls for sensor maintenance or complete structural change. It appears that a practical solution of performing bias update would be sufficient. We will revisit and analyze this issue in the following section in order to confirm whether the operating conditions change so dramatically that one would need to change the soft sensor structure.

\subsection{Design of Inferential Sensors for the FCC Unit using Randomly Distributed Data}\label{sec:res_fcc_rand}
We randomly distribute 50\,\% of the available data to the training set and the remaining data to the testing set. We generate 50 such distributions to increase the interpretabilily of the results. We then use the same workflow to design the soft sensors as outlined above.

\begin{table}
	\caption{Comparison of the number of inputs $n_\text{p}^*$ (number of principal components of PCA and PLS), sensor accuracy (RMSE) and bias correction relative frequency (BC) over 50 random training/testing data distributions for the FCC unit.}
	\label{tab:fcc_res_rand}
	\centering
	\begin{tabular}{c|c c c c c c|c}
		\toprule
		         & OLSR  & PCA     & PLS     & LASSO & SS    & SS-CV & Ref   \\ \midrule
		$n_\text{p}^*$  & 11    & 11\,(7) & 11\,(8) & 7     & 6     & 5     & 3     \\
		RMSE     & 0.105 & 0.104   & 0.106   & 0.106 & 0.106 & 0.110  & 0.121 \\
		BC [\%]  & 23.0 & 24.3   & 23.0   & 24.3 & 27.0 & 24.3 & 28.4 \\
		\bottomrule
	\end{tabular}
\end{table}

We report averages of $n_\text{p}^*$, RMSE and BC for each soft sensor over the 50 data distributions in Table~\ref{tab:fcc_res_rand}. According to the sensor complexity criterion ($n_\text{p}^*$), we can see that the designed soft sensors suggest more complex structure (at least two extra input variables) compared to the reference structure and also compared to the previous scenario with chronological training/testing data assignment. This suggests that varying operating conditions in the plant would require frequent revision of the sensor structure for better performance. The performance of the designed advanced sensors does not improve compared to the designs using chronological training/testing data distribution, which is a consequence of the overfitting implied by the increased complexity of the sensor. For example, LASSO and both SS methods commonly suggest including $T_\text D$ and $Q_\text{B}$ on top of the inputs suggested in the previous section. However, none of these variables seem to be significantly useful for the sensor overall. While, unlike for distillate temperature $T_\text D$, inclusion of $Q_\text{B}$ would make sense from process viewpoint, its effect is already present in the input $Q_\text B/F$. Only the inferential sensor designed by OLSR exhibits improved accuracy compared to the design with chronologically distributed training/testing data. This is a consequence of providing better training data (more similar to testing ones) to the sensors, which reduces the overfitting effect. Designed advanced soft sensor (including PCA) shows the increased frequency of the bias correction, which can be attributed to the large noise magnitude in the lab data and overfitting.

The performance features of the particular sensors remain practically the same as in the case of chronological training/testing data distribution. The soft sensor designed by PCA is slightly more accurate than other soft sensors and it improves the accuracy of the reference soft sensor by about 14\,\%. Yet the drop in this improvement confirms the overfitting. The structure of the soft sensor designed by SS-CV is less complicated than the structures of other designed soft sensors. As expected, the least complex sensor designed by SS-CV is again followed in terms of performance by design using the SS and LASSO methods, respectively.

\begin{figure}
	\centering
	\includegraphics[width=\linewidth]{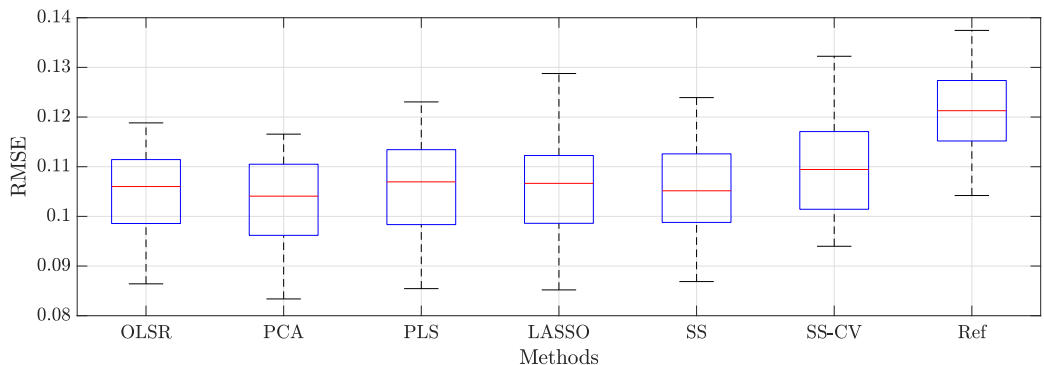}
	\caption{Comparison of accuracy of the designed inferential sensors over	
	50 different random training/testing data distributions for the FCC unit.}
	\label{fig:fcc_rand_IS_rmse}
\end{figure} 

Figure~\ref{fig:fcc_rand_IS_rmse} visualizes the accuracy statistics of each soft sensor from the 50 randomly generated training/testing datasets using box plots. The central horizontal-line marker indicates the median, the bottom and top edges of the box indicate the $25^\text{th}$ and $75^\text{th}$ percentiles, respectively, the whiskers extend to the most extreme data points not considered outliers, and the outliers are plotted individually using the '+' symbol. We can see that the median performance mostly copies the average performance of the designed soft sensors outlined in Table~\ref{tab:fcc_res_rand}.

The accuracy variance seems to be considerable for all sensors, which confirms the aforementioned large noise in the samples the possible sensor overfitting. The least variance is present in the reference sensor, which is due to the aforementioned robustness properties.

\begin{figure}
	\centering
	\begin{subfigure}{\textwidth}
		\includegraphics[width=\linewidth]{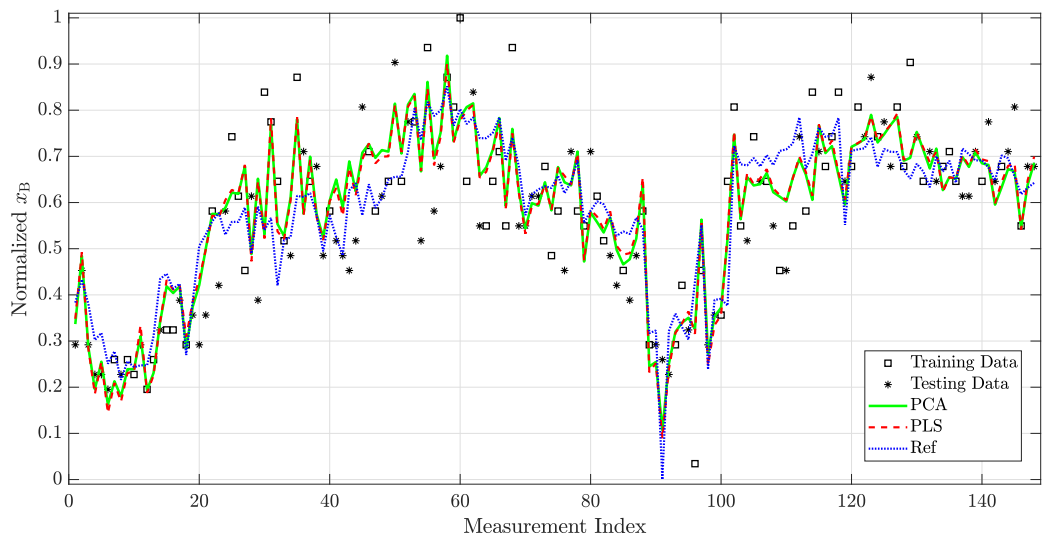}
		\caption{Training and prediction performance of the reference sensor and of the sensors designed by PCA and PLS methods.}
		\label{fig:fcc_rand_IS_PCA_PLS}
	\end{subfigure}
	\begin{subfigure}{\textwidth}
		\vspace{.5cm}
		\includegraphics[width=\linewidth]{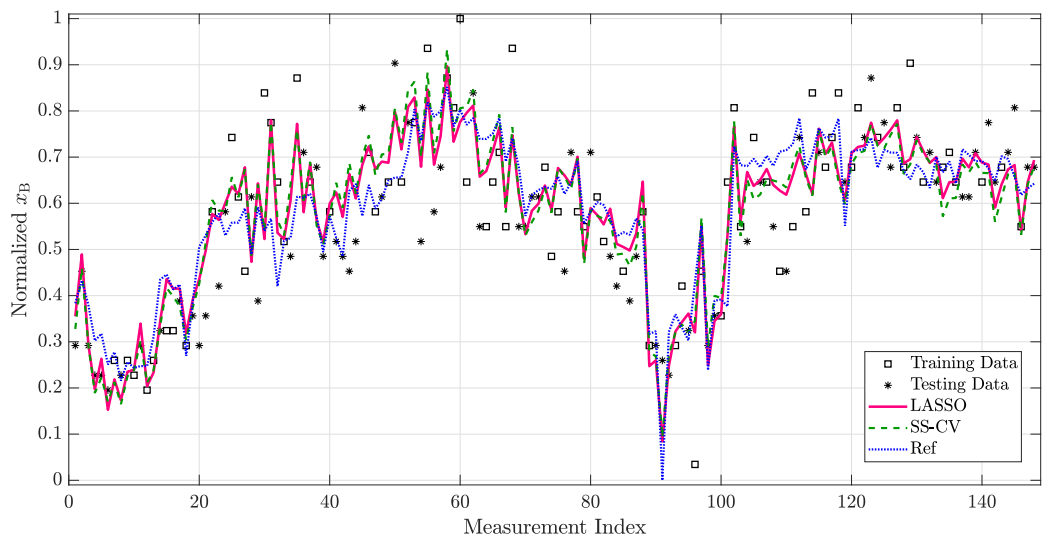}
		\caption{Training and prediction performance of the reference sensor and of the sensors designed by LASSO and SS-CV methods.}
		\label{fig:fcc_rand_IS_LASSO_SS}
	\end{subfigure}
	\caption{Comparison of the soft sensors using randomly distributed training/testing data for the FCC unit.}
	\label{fig:fcc_rand_IS}
	\vspace{0cm}
\end{figure}

As in the previous section, we visualize the training and prediction performance of the designed soft sensors in Figure~\ref{fig:fcc_rand_IS} for one representative random training/testing data distribution. We again show results obtained for the reference soft sensor (both plots; blue dotted line),  the soft sensors designed by PCA and PLS (Figure~\ref{fig:fcc_rand_IS_PCA_PLS}; green solid line and red dashed line, respectively), and the soft sensors designed by LASSO and SS-CV (Figure~\ref{fig:fcc_rand_IS_LASSO_SS}; magenta solid line and green dashed lines, respectively).

As can be expected, the performance of the soft sensors is similar to the performance of the soft sensors designed by using time series data. The previously discussed discrepancy between the sensors and the measurements (around the measurements 130--148) is decreased. This, together with the increased complexity of the sensors designed using randomly distributed data, leads us to the conclusion that the performance of an advanced sensor can only be maintained if the sensor structure changes frequently or if the sensor parameters are frequently updated. Of course, in this particular case, the problem would be practically resolved by bias update.

\subsection{Inferential Sensors for the VGH Unit}
The available historical data encompasses almost two years of production in the period 2018--2019 with 34,845 time points of online measurements. This is a comparable amount of data as in the previous case study. The desired output variable $T_{95\%,\text{HGO}}$, which, similarly to the previous use case, indicates the purity of the distillation product, is determined by the lab analysis. The mentioned time span involves 689 measurements of the output variable as it is measured more frequently than in the case of FCC.

\begin{figure}
	\centering
	\captionsetup[subfigure]{labelfont=bf}
	\begin{subfigure}{\textwidth}
		\includegraphics[width=\linewidth]{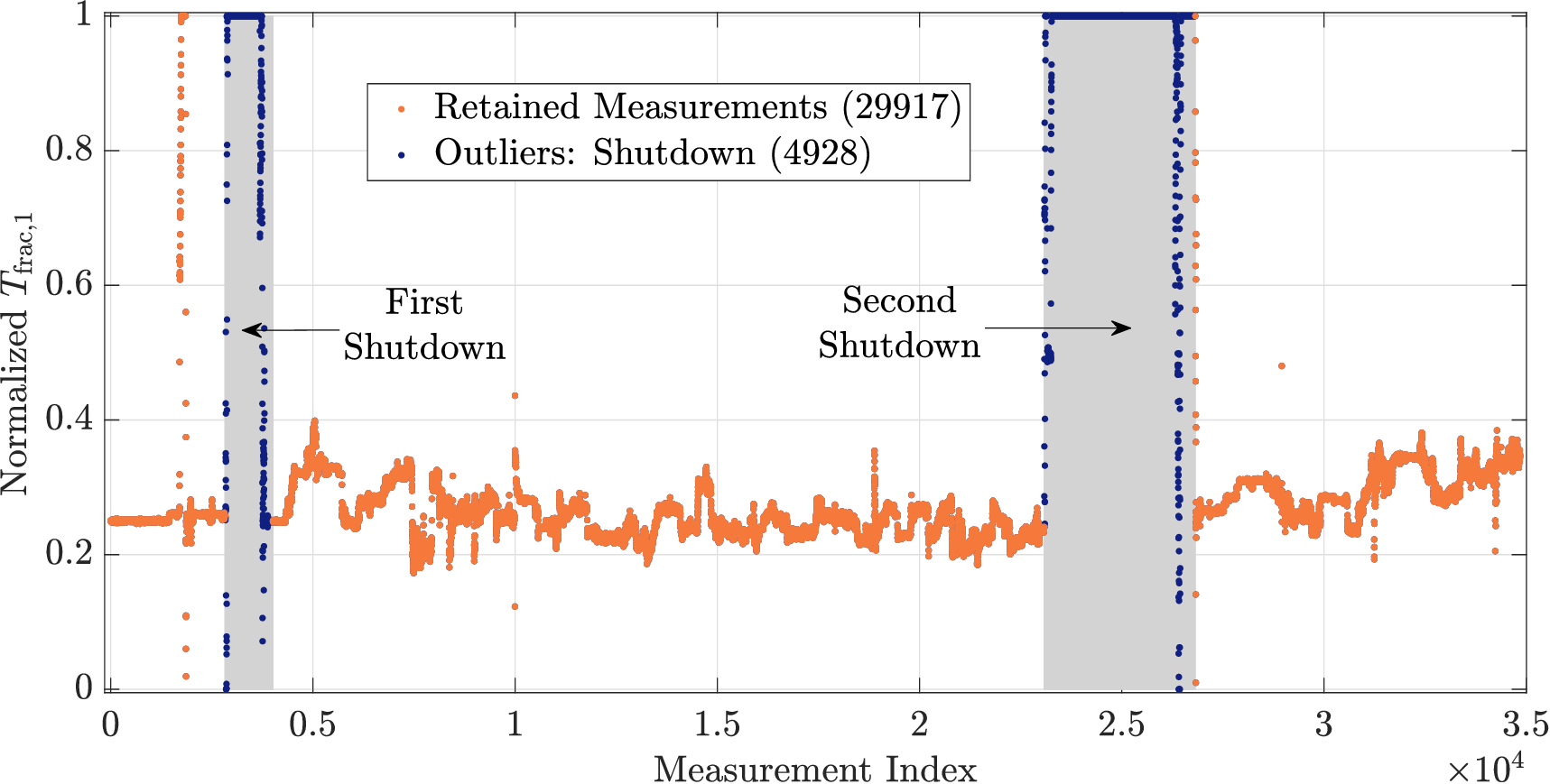}
		\caption{Data pre-treatment by visual inspection detecting plant shutdowns.}
		\label{fig:vgh_GED_shutdown}
	\end{subfigure}
	\begin{subfigure}{\textwidth}
		\vspace{.5cm}\hspace{-1cm}
		\includegraphics[width=1.1\linewidth]{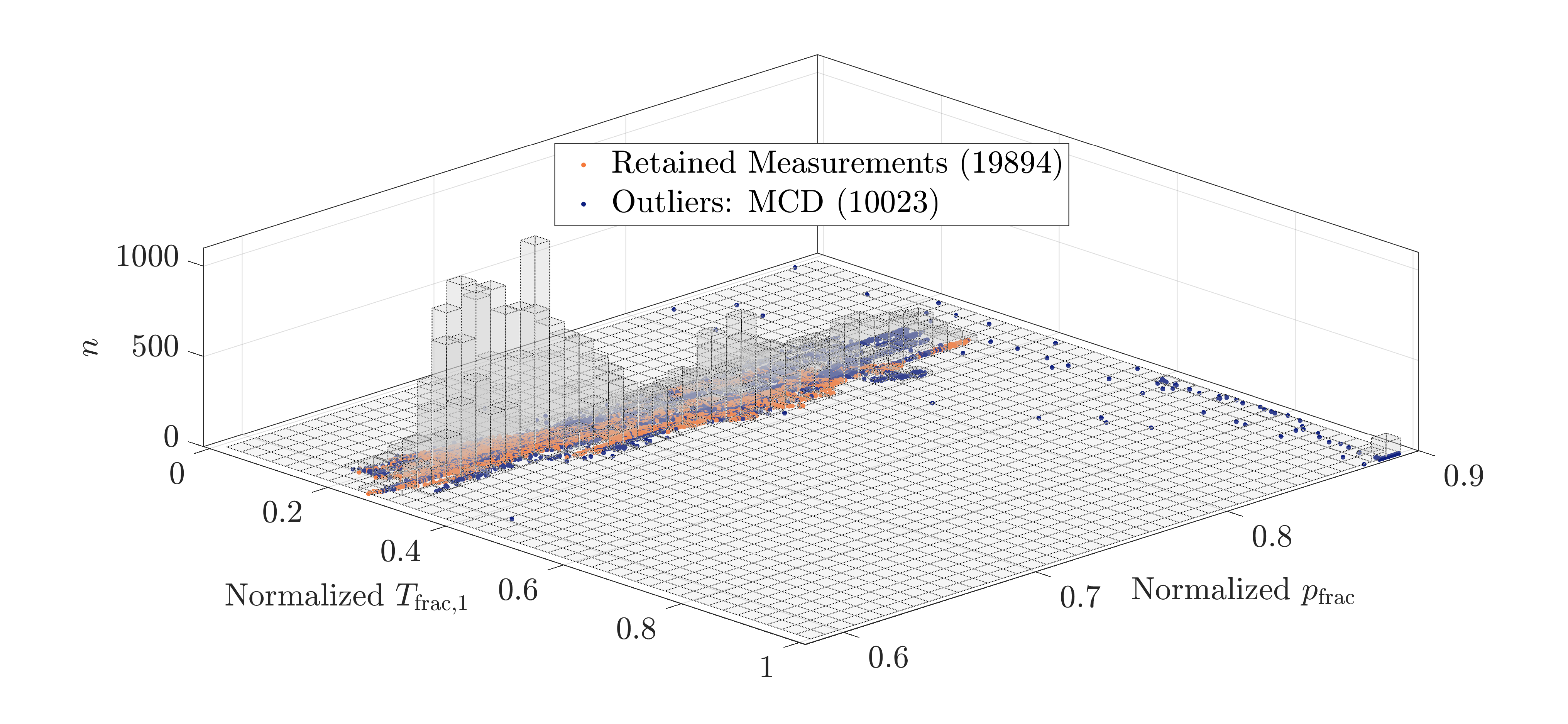}
		\caption{Data treatment by the MCD method.}
		\label{fig:vgh_GED_MCD_hist}
	\end{subfigure}
	\caption[]{\footnotesize \textbf{(a)} Normalized temperature in the main fractionator of the VGH unit vs.~measurement index. \textbf{(b)} Histogram of the temperature vs.~pressure in the main fractionator of the VGH unit and retained measurements vs.~outliers after data treatment.}
	\label{fig:vgh_GED}
\end{figure}

We first perform the pre-treatment of the available data. Based on the visual inspection of the time series of a temperature in the main fractionator (Figure~\ref{fig:vgh_GED_shutdown}), we eliminated two intervals with obviously deviated measurements (see gray intervals in Figure~\ref{fig:vgh_GED_shutdown}). The unit operators confirmed that the omitted 4,928 measurement points (black points in Figure~\ref{fig:vgh_GED_shutdown}) correspond to the unit shutdowns.

Subsequently, the remaining data is processed using the $T^2$ distance, MCD, and $k$-means clustering methods. Although the visualized temperature data does not seem to be much qualitatively different in nature than the case of the FCC unit, there are more distinct variations and steady states. This feature causes that the $T^2$ distance method suggests removing more outliers than MCD and $k$-means clustering.

For the case of the $T^2$ distance method, 13,874 outliers is indicated, which represents almost half of the available pre-treated data. This behavior can be attributed to the previously observed distinct variations and steady states, which bias the statistics used in the $T^2$ distance method. In fact, if we wanted to tune this method to the similar performance as the MCD method, we would require increasing the probability of measurements acceptance from 99.7\,\% to 99.9\,\%. This seemingly small alteration represents a significant increase in the acceptance, by one half of a standard deviation. 

The MCD method indicates slightly more outliers (10,023 measurements) as in the case of the FCC unit (6,917 measurements), which may be caused by the worse quality of the data from the VGH unit. The $k$-means clustering method indicates much more outliers (11,229 measurements) than in the FCC unit (265 measurements). It uses 21 clusters (compared to five clusters detected for the FCC unit), which seems to be a consequence of the distinct variations and steady states. Nonetheless, the data distribution among the clusters exhibits certain uniformity, which further demonstrates the sensitivity of the $k$-means clustering method to tuning (e.g., number of clusters). 

\begin{figure}
	\centering
	\includegraphics[width=\linewidth]{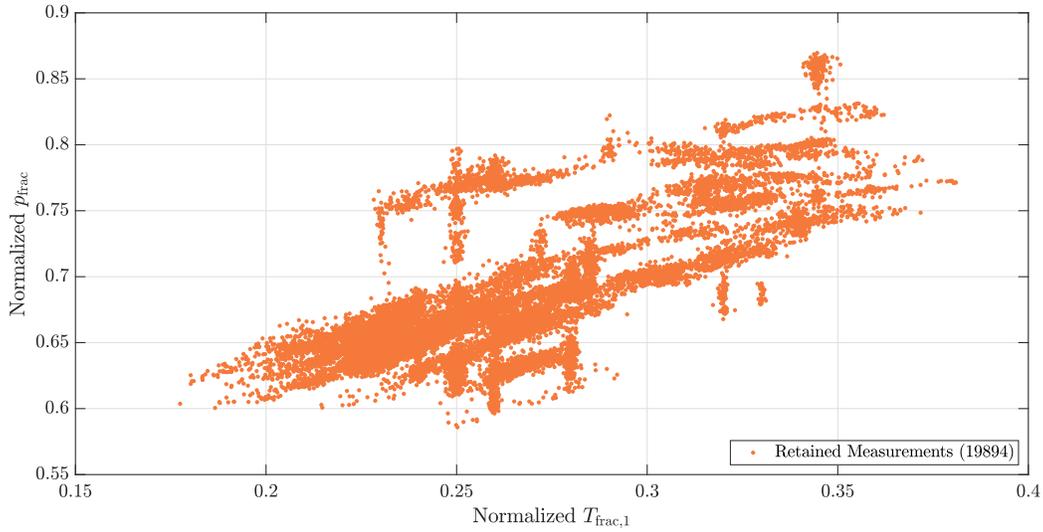}
	\caption{The retained online measurements (by the MCD method) of the temperature and pressure in the main factionator of the VGH unit.}
	\label{fig:vgh_GED_vars_without_GE}
\end{figure}

As in the case of the FCC unit, we again choose to remove the outliers labeled by the MCD method as it retains reasonable amount of data points. Even though the data quality (e.g., number of shutdowns, variations of the operating conditions) of the VGH unit is worse than the FCC unit, we can see more minor differences among the applied data-treatment methods. Therefore, only the performance of the MCD method is further shown via the histogram of data points of temperature vs.~pressure in the main fractionator in Figure~\ref{fig:vgh_GED_MCD_hist}. The blue points represent indicated outliers and the rest of the data (green points) is retained for the design of soft sensors (Figure~\ref{fig:vgh_GED_vars_without_GE}). We can conclude that the marked outliers are mostly measurements deviated from the area with the highest density of the measurements. This proves the effectiveness of the MCD method to indicate deviated and undesirable measurements.

\subsection{Design of Inferential Sensors for the VGH Unit using Time Series Data}\label{sec:res_vgh_chron}
We design inferential sensors in the same way as in Section~\ref{sec:res_fcc_chron}. Therefore, we distribute (chronologically) first 50\,\% of the available time series data to training set and last 50\,\% of available time series data to the testing set.

Soft-sensors designed by PCA and PLS require thirteen and fifteen principal components, respectively, to explain 98\,\% of the variance in the data. This, on the one hand, suggests possible overfitting yet, on the other hand, there seems to be a good agreement between the advanced design methods on the number of important variables (or their combinations), i.e., 13--15. When designing a soft sensor by the SS methods, similar to the previous use case, we found that the combination with the $\mathit{R}^2_{\text{adj}}$ criterion gave the best results. Unlike in the case of the FCC unit, the SS-CV method proposes different sensor structure as the SS method (with $\mathit{R}^2_{\text{adj}}$ criterion).

\begin{table}
	\caption{Comparison of the number of inputs $n_\text{p}^*$ (number of principal components of PCA and PLS), sensor accuracy (RMSE) and bias correction relative frequency (BC) using time series data for the VGH unit.}
	\label{tab:vgh_res_chron}
	\centering
	\begin{tabular}{c|c c c c c c|c}
		\toprule
		         & OLSR  & PCA      & PLS      & LASSO & SS    & SS-CV & Ref   \\ \midrule
		$n_\text{p}^*$  & 19    & 24\,(13) & 22\,(15) & 14    & 15    & 12    & 1     \\
		RMSE     & 0.184 & 0.103    & 0.158    & 0.145 & 0.190  & 0.182 & 0.114 \\
		BC [\%]  & 82.8 & 85.3    & 80.2    & 74.6 & 82.3 & 79.3 & 75.4 \\
		\bottomrule
	\end{tabular}
\end{table}

A comparison of the designed sensors in terms of their complexity ($n_\text{p}^*$), accuracy (RMSE), and the amount of bias correction (BC) is shown in Table~\ref{tab:vgh_res_chron}. As we can see, the suggested structure ($n_\text{p}^*$) of the designed soft sensors is much more complicated than the structure of the reference soft sensor. Out of 30 candidate inputs, the designed sensors suggest to include at least eleven more inputs. All the design methods (even OLSR) are able to sparsify to a certain extent the structure of the full sensor~\eqref{eq:full_sensor_vgh}. Beside $PCT_{\text{HGO}}$ included in the reference sensor, LASSO suggests involving $T_{\text{frac,2}}$, $T_{\text{frac,1}}$, $T_{\text{f,50p}}$ and $x_{\text{H}2}$ among the most influential variables. On contrary, the SS methods suggest including $T_{\text{frac,2}}$, $T_{\text{frac,1}}$, $PCT_{\text{GF}}$ and $T_{\text{wabt,1}}$. Despite the disagreement on the added variables, it appears that certain variables from the reaction section of the plant could play a role in explaining the bad performance of the reference sensor when qualitatively different feedstock is used.

Overall, the accuracy of the designed soft sensors (see RMSE in Table~\ref{tab:vgh_res_chron}) shows the best performance for the soft sensor designed by PCA, good performance of the soft sensors designed by the PLS and LASSO methods and the worst performance of the soft sensors designed by OLSR and SS methods. Apparently, the reference sensor shows high robustness. The poor accuracy of the soft sensor designed by SS methods can be explained by the highly varying operating conditions of the plant. This can also be documented by the much increased amount of bias correction compared to the case of the FCC unit (see in Table~\ref{tab:fcc_res_chron}).

We can see that the soft sensor designed by OLSR is much more complicated, less accurate and more frequently corrected than the reference soft sensor. The results show that PCA and PLS methods are not able to reduce the dimensionality of the soft sensor compared to OLSR. The high number of principal components of these methods also suggests that a complex structure is required to express the behavior of the desired variable. The soft sensors designed by PCA, PLS and LASSO are more accurate than other designed soft sensors. Nevertheless, only the PCA sensor is more accurate (by about 10\,\%) than the reference soft sensor. According to the values of the BC criterion in Table~\ref{tab:vgh_res_chron}, the soft sensor designed by PLS is more appropriate than the PCA soft sensor, although both sensors are more frequently corrected than the reference soft sensor. The further values of BC indicate that soft sensors designed by LASSO and SS-CV are corrected less frequently than other designed soft sensors, which results from their simple structure (and implied robustness).

\begin{figure}
	\centering
	\begin{subfigure}{\textwidth}
		\includegraphics[width=\linewidth]{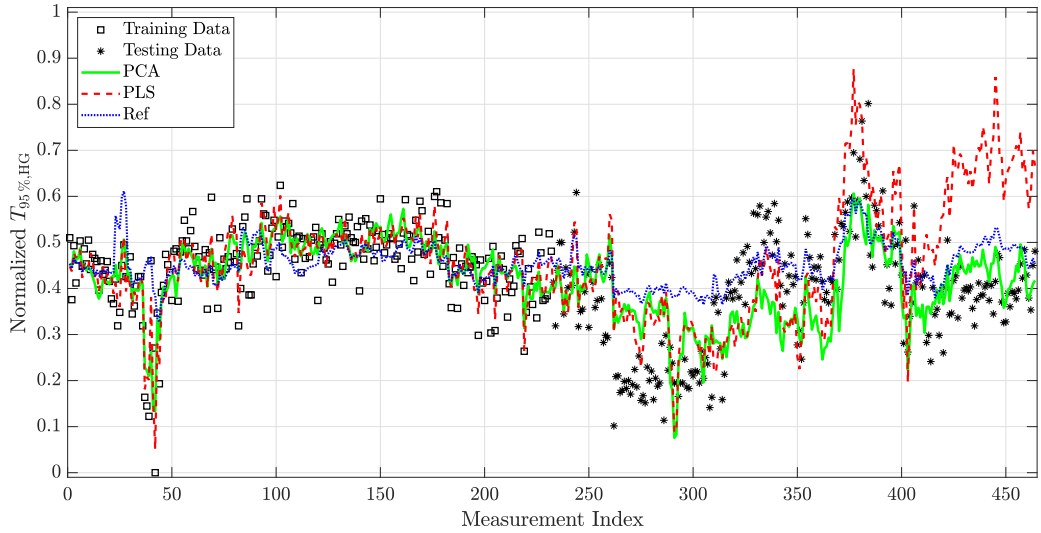}
		\caption{Training and prediction performance of the sensors designed by PCA and PLS methods and reference sensor.}
		\label{fig:vgh_chron_IS_PCA_PLS}
	\end{subfigure}
	\begin{subfigure}{\textwidth}
		\vspace{.5cm}
		\includegraphics[width=\linewidth]{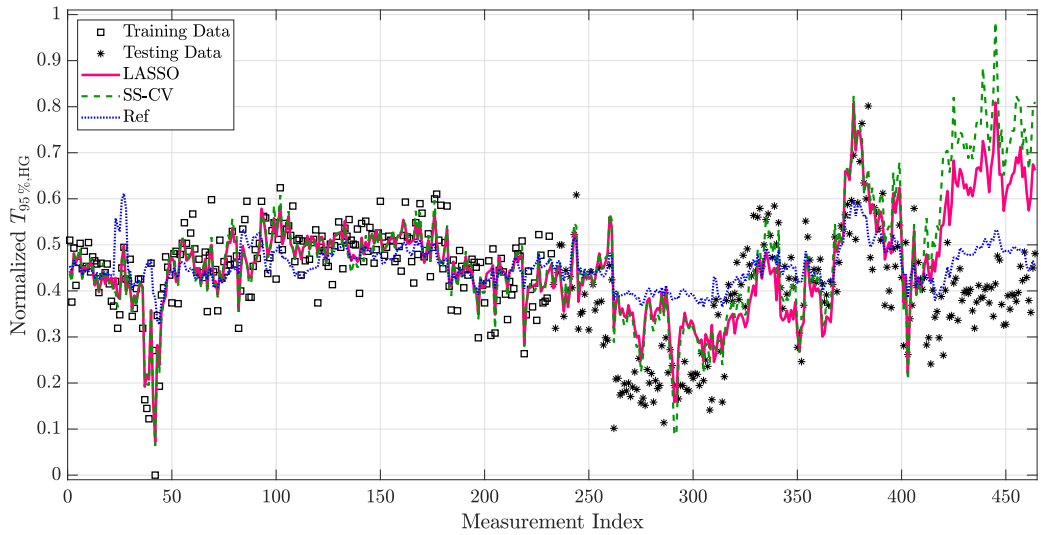}
		\caption{Training and prediction performance of the sensors designed by LASSO and SS-CV methods and reference sensor.}
		\label{fig:vgh_chron_IS_LASSO_SS}
	\end{subfigure}
	\caption{Comparison of the soft sensors for the VGH unit designed using time series data.}
	\label{fig:vgh_chron_IS}
	\vspace{0cm}
\end{figure}

In order to provide more comprehensive comparison of the soft sensors, we visualize their predictive performance on the output variable in Figure~\ref{fig:vgh_chron_IS} using the same color coding as in the previous case study. The data shows high variability indicating several changes of the operating conditions within the studied time window, in both training and testing datasets. Nonetheless, the variability within the testing set appears to be higher. This might explain the poor performance of the designed advanced sensors and it is confirmed by the high effort of bias correction.

Figure~\ref{fig:vgh_chron_IS_PCA_PLS} further presents the training and testing (predictions) performance of the designed advanced soft sensors, by PCA and PLS (Figure~\ref{fig:vgh_chron_IS_PCA_PLS}) and by LASSO and SS-CV (Figure~\ref{fig:fcc_chron_IS_LASSO_SS}), compared to the reference soft sensor. We can directly see the training performance of the designed advanced inferential sensors being much better than the reference soft sensor. However, there are several sections in the testing dataset, where these soft sensors are not able to explain the behavior of the output variable. This is most prominent around the measurements 260--320 and 420--464. Interestingly, PCA-based soft-sensor performs relatively well in both the designated periods, which suggests that some process features were successfully caught in the sensor. On the other hand, it exhibits a relatively poor performance around measurement index 350, where it is outperformed by other sensors (even the reference sensor). These observations suggest that the training set is poor and should be expanded. 

It appears that a practical solution of performing bias update would be sufficient in this situation. We will revisit and analyze this situation in the following section in order to confirm whether the operating conditions change so dramatically that one would need to vary the soft-sensor structure often.

\subsection{Design of Inferential Sensors for the VGH Unit using Randomly Distributed Data}\label{sec:res_vgh_rand}
Next, we design soft sensors using randomly distributed data. We assign 50\,\% of the available randomly distributed data to the training set and the remaining data to the testing set. We generate 50 such distributions and we use the same training/testing workflow as above. We finally present the average performance measures from the different runs of the corresponding soft-sensor design.

\begin{table}
	\caption{Comparison of the number of inputs $n_\text{p}^*$ (number of principal components of PCA and PLS), sensor accuracy (RMSE) and bias correction relative frequency (BC) over 50 random training/testing data distributions for the VGH unit.}
	\label{tab:vgh_res_rand}
	\centering
	\begin{tabular}{c|c c c c c c|c}
		\toprule
		         & OLSR  & PCA      & PLS      & LASSO & SS    & SS-CV & Ref   \\ \midrule
		$n_\text{p}^*$  & 24    & 25\,(15) & 25\,(17) & 15    & 16    & 12    & 1     \\
		RMSE     & 0.086 & 0.087    & 0.085    & 0.086 & 0.086 & 0.087 & 0.105 \\
		BC [\%]  & 88.8 & 86.6    & 86.6    & 90.1 & 89.7 & 87.1 & 91.0 \\
		\bottomrule
	\end{tabular}
\end{table}

The comparison of soft sensors in Table~\ref{tab:vgh_res_rand} involves the same criteria ($n_\text{p}^*$, RMSE, BC) as in the previous section. In terms of complexity of the designed sensors, we see similar trend as in the FCC use case. The overall complexity of the designed soft sensors is mostly higher compared to the soft sensors designed on chronologically distributed data (see Table~\ref{tab:vgh_res_chron}). This is a recurring observation (from the first case study) and points at the need of enriching the number of explaining variables to adapt for varying plant operating conditions. Only the soft sensor designed by SS-CV is an exception and it even maintains exactly the same sensor structure. These observations reveal that despite SS-CV found good sensor structure in case of chronologically distributed data, the variation in the operating conditions would require to adapt sensor parameters. This also definitely proves high variation of operating conditions and its strong influence on the sensor performance. Similar to SS-CV, also for the rest of the designed sensors the most influential inputs selected by the design methods remain unchanged compared to the case of chronological training/testing data distribution. Each designed soft sensor shows the increased frequency of the bias correction, which can be attributed to the large noise magnitude in the lab data and to the need for adapting the sensor frequently due to operating conditions.

The accuracy of the designed inferential sensors is essentially the same and each sensor is more accurate than the reference sensor. The most accurate sensor is designed by PLS and it improves the accuracy of reference sensor by about 19\,\%. A drop in this performance by PCA-based sensor can be attributed to significant changes in the operating conditions in combination with changes in the sensitivity of the output variable to different inputs (online measurements). The latter claim is supported by the comparatively better performance of the sensor designed by PLS.

\begin{figure}
	\centering
	\includegraphics[width=\linewidth]{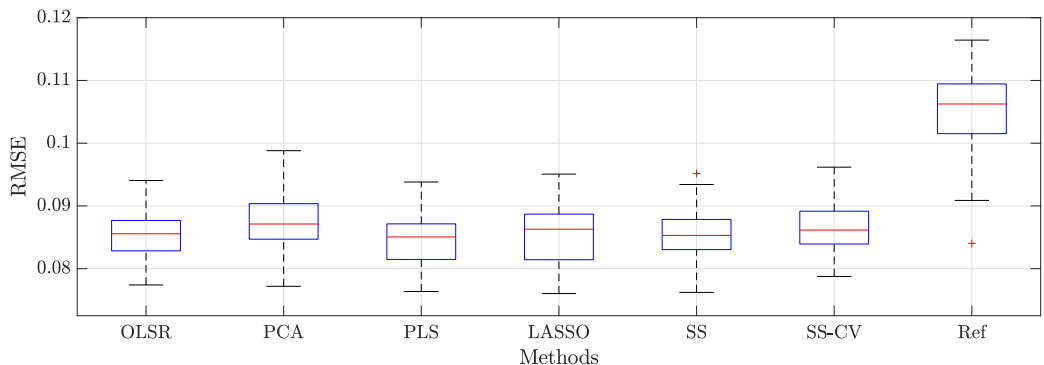}
	\caption{Comparison of the designed inferential sensor accuracy (RMSE) over 50 different randomly generated training/testing data distributions for the VGH unit.}
	\label{fig:vgh_rand_IS_rmse}
\end{figure} 

Figure~\ref{fig:vgh_rand_IS_rmse} visualizes the accuracy statistics using box plots of each soft sensor from the 50 randomly distributed training/testing datasets. We can see that the performance statistics of all the designed soft sensors mostly copies the conclusions reached in the discussion on the average performance (see Table~\ref{tab:vgh_res_rand}). The results show similar accuracy variance of each soft sensor, which means that the variance is caused mainly by the particular noise realizations in the data. The smallest variance is though achieved for the sensors found by SS methods.

\begin{figure}
	\centering
	\begin{subfigure}{\textwidth}
		\includegraphics[width=\linewidth]{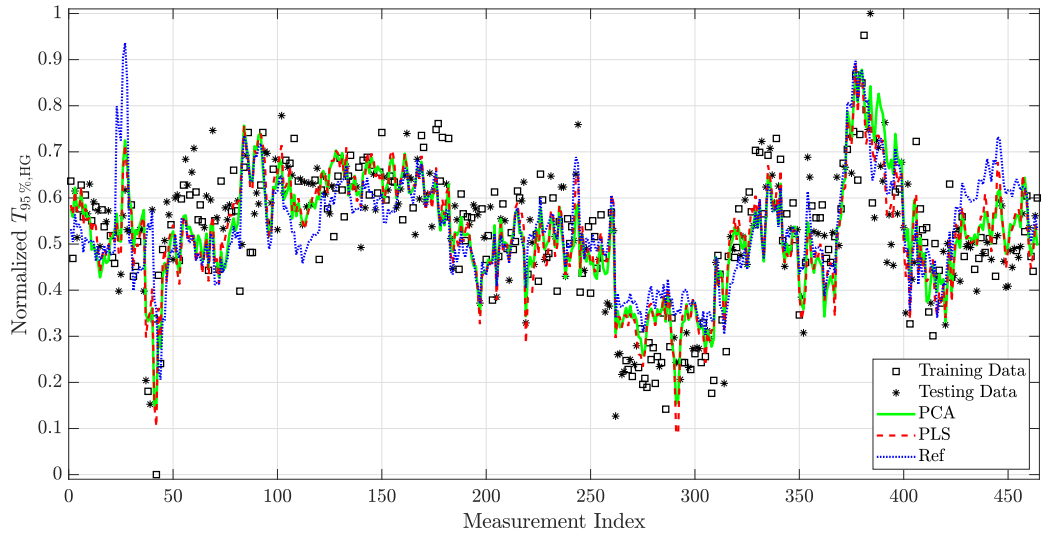}
		\caption{Training and prediction performance of the sensors designed by PCA and PLS methods and reference sensor.}
		\label{fig:vgh_rand_IS_PCA_PLS}
	\end{subfigure}
	\begin{subfigure}{\textwidth}
		\vspace{.5cm}
		\includegraphics[width=\linewidth]{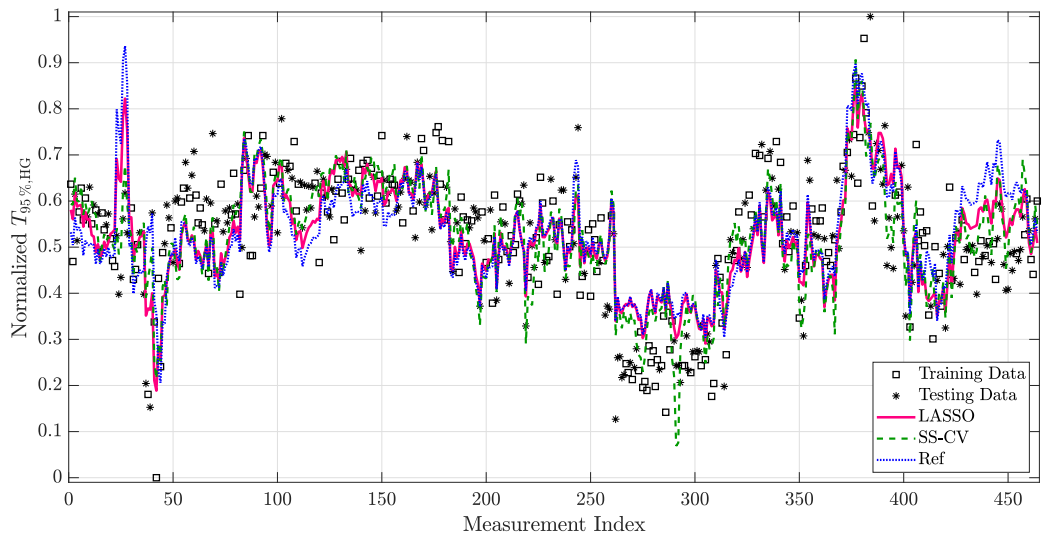}
		\caption{Training and prediction performance of the sensors designed by LASSO and SS-CV methods and reference sensor.}
		\label{fig:vgh_rand_IS_LASSO_SS}
	\end{subfigure}
	\caption{Comparison of the soft sensors using randomly generated training/testing data distribution for the VGH unit.}
	\label{fig:vgh_rand_IS}
	\vspace{0cm}
\end{figure}

As in the previous section, Figure~\ref{fig:vgh_rand_IS} visualizes the training and prediction performance of the designed soft sensors for one representative random training/testing data distribution. Results are shown for the reference soft sensor (both plots), the soft sensors designed by PCA and PLS (Figure~\ref{fig:vgh_rand_IS_PCA_PLS}), and the soft sensors designed by LASSO and SS-CV (Figure~\ref{fig:vgh_rand_IS_LASSO_SS}). The performance improvement of the soft-sensors with randomly distributed data compared to chronological data is evident.  We can observe this on previously mismatched measurements around markers 420--464. Yet, we can clearly identify the period of measurements 260--320 that still exhibits unsatisfactory sensor performance. This calls for another investigation at the plant and revision of the set of candidate sensor inputs.

In conclusion, the advanced design methods show great potential for improving the sensor accuracy beside the good robustness properties of the reference sensor. Yet due to the complexity of the use case, the price to pay for the improved performance is paid in terms of higher sensor complexity. Moreover, due to varying operating conditions, the advanced sensors would need to be often updated or trained on a carefully selected training set.

\section{Discussion}\label{sec:discussion}
Overall, we can say that each studied data treatment method is able to a certain extent indicate outliers in the multivariate data. The advantage of the $T^2$ distance method is a simple principle. However, this method is strongly affected by the number of treated variables or by the data distribution. The $T^2$ distance method selects fewer outliers in the FCC unit data (3,567 outliers) than the VGH unit (13,874 outliers). The best results were achieved using the MCD method, which guarantees higher quality of the retained data than the $T^2$ distance method. The performance of this method seems to be consistent in both case studies. The MCD method indicates 6,917 outliers in the FCC unit and 10,023 outliers in the case of the VGH unit. The higher number of indicated outliers in the case of the VGH unit is caused mainly by the worse quality of the data. The treatment of the industrial data pointed out that $k$-means clustering is quite sensitive to tuning (e.g., number of clusters) that might lead to inferior-quality data treatment. We can see an even more significant discrepancy between the number of indicated outliers in the FCC unit and VGH unit (265 outliers and 11,229 outliers, respectively) by $k$-means clustering as in the $T^2$ distance method. It seems that the performance of this method should be adjusted to select more outliers in measurements in the case study on the FCC unit.

The performance of the inferential sensors designed by the studied data-based method (OLSR, PCA, PLS, LASSO, SS and SS-CV) is compared against the reference (current) sensor in both case studies. The reference sensor has a relatively simple structure (three input variables) in the FCC unit and a simple structure (one input variable) in the VGH unit. The low structural complexity provides higher robustness of the inferential sensors. We could see this robustness when the inferential sensors were designed according to the chronological training/testing dataset of the VGH unit. In this case, the designed advanced inferential sensors are more complex yet less accurate than the reference sensor in the final section of the testing dataset. It is most likely that the process deviates from the operating conditions present during the training phase and the advanced sensors would require frequent parameter adaptation to maintain the designed performance.

The results from chronological distribution of training/testing dataset indicate that the inferential sensor designed by PCA achieved the highest accuracy. It outperforms the reference sensor by about 18\,\% in the FCC unit and by about 10\,\% in the VGH unit. Such sensor could be used for plant monitoring. On the other hand, if we also consider the sensor complexity, then the SS-CV method outperforms the rest of the approaches. A low-complexity sensor would be more suitable for optimization or advanced control.

The design of inferential sensors considering both chronologically and randomly distributed training/testing datasets seems to be an effective way to determine the impact of changing operating conditions in the process. The results suggest that the inferential sensors designed over the chronologically distributed training/testing dataset are less sensitive to overfitting than the randomly distributed training/testing dataset. This phenomenon supports the hypothesis of the occurrence of varying operating conditions since the trained sensors tend to involve more inputs to model the changing conditions.

Our investigation has also found that inferential sensors commonly used in the petrochemical industry show high robustness and can give solid performance even long after their commissioning. On the other hand, the relative simplicity of the structure can be easily enhanced in simple cases (the FCC unit use case) by extension of the structure without much maintenance effort. Such sensors can also improve the trust of the operators in the sensors and the automation technology. For this purpose, advanced methods of soft-sensor design (LASSO and SS methods) show a good promise and even the associated computational burden is justified. In more complex cases, the studied design methods can be a promising technology for root-cause analysis.

\section{Conclusions}\label{sec:conclusions}
This paper studied soft (inferential) sensors design to monitor unmeasurable variables in the petrochemical industry. Due to the presence of systematic errors and outliers in the industrial measurements, some well-known data pre-treatment methods ($T^2$ distance, MCD, $k$-means clustering) were used and compared. The results suggest that MCD is more versatile than $T^2$ distance or $k$-means clustering and it performs well overall. Furthermore, the data quality seems to be very well reflected by the number of indicated outliers by the MCD method.

The data retained after the treatment by MCD was subsequently used to design inferential sensors using several data-based methods (OLSR, PCA, PLS, LASSO, SS and SS-CV). The results indicate that PCA tends to design more accurate yet more complex inferential sensors than other methods. On the other hand, the SS-CV method provides well-performing yet structurally less complex sensors than other methods. Therefore, this method is recommended for the design of as simple inferential sensor as possible. A good compromise between the accuracy and complexity is represented LASSO and SS (with overfitting criteria).

The results also indicate that the designed inferential sensors cannot predict the desired variable behaviour over a long time span. There often occur data sections where the designed soft sensors significantly deviate from the measurements of the desired variable. The bias correction seems to be an effective and simple remedy for these discrepancies. In our future work, we will concentrate on finding effective methods of sensor adaptation and/or an efficient way of combining several inferential sensors to cover more operating conditions in the industrial unit.

\section*{Acknowledgments}\label{sec:acknowledgments}
This research is funded by the Slovak Research and Development Agency
under the projects 15-0007, 20-0261, and SK-FR-2019-0004, and by the
Scientific Grant Agency of the Slovak Republic under the grant
1/0691/21. MM, MF, and RP would like to thank Dr. Andr\'as Buti from
Slovnaft, a.s in Bratislava for several inspiring discussions and for
his fruitful comments on the presented work.

\bibliographystyle{elsarticle-harv}
\bibliography{references}
\end{document}